\documentclass[10pt, onecolumn,letterpaper]{article}

\usepackage[pagenumbers]{cvpr} 

\newcommand{\thename}[0]{RAD-2}

\usepackage{pifont}
\usepackage{multirow}
\usepackage[table]{xcolor}
\usepackage{makecell}
\usepackage{bbm}
\usepackage{marvosym} 
\usepackage{array}
\usepackage{makecell}
\usepackage{colortbl}
\usepackage{sectsty}
\usepackage{enumitem}

\definecolor{horizonblue}{RGB}{0, 102, 204}

\definecolor{cvprblue}{rgb}{0.21,0.49,0.74}
\definecolor{ourlightblue}{RGB}{245,247,255}
\usepackage[pagebackref,breaklinks,colorlinks,allcolors=horizonblue]{hyperref}
\def\thename{InfiniteVL}

\def\paperID{*****} 
\def\confName{CVPR}
\def\confYear{2026}

\definecolor{horizonblue}{RGB}{0, 102, 204} 

\usepackage{graphicx}
\usepackage{booktabs}
\usepackage{siunitx}
\usepackage{makecell}
\usepackage{array} 
\usepackage{multirow}
\usepackage{multicol}
\usepackage{color,xcolor}
\usepackage{colortbl}
\usepackage{graphicx}
\usepackage{caption}
\usepackage{seqsplit}  

\usepackage{tabularx,booktabs,makecell,array, adjustbox, changepage}

\usepackage[accsupp]{axessibility}  

\usepackage{hyperref}

\usepackage{orcidlink}

\usepackage[many]{tcolorbox}

\usepackage{titlesec}
\titleformat*{\section}{\large\bfseries\color{horizonblue}}

\makeatletter
\renewenvironment{abstract}{%
    \vspace{-1.0em} 
    \begin{tcolorbox}[
        enhanced,
        colback=white,           
        colframe=horizonblue,    
        boxrule=1pt,             
        arc=8pt,                 
        outer arc=8pt,
        left=15pt, right=15pt, top=8pt, bottom=12pt 
    ]
    \begin{center}
        {\centering \large\bfseries\color{horizonblue} Abstract}
    \end{center}
    \vspace{0em} 
    \itshape\ignorespaces 
}{%
    \end{tcolorbox}
    \vspace{2em} 
}
\makeatother

\title{
    \parbox{\textwidth}{
        \vspace*{-3em} 
        \noindent
        \makebox[\textwidth][s]{ 
            \includegraphics[height=2em]{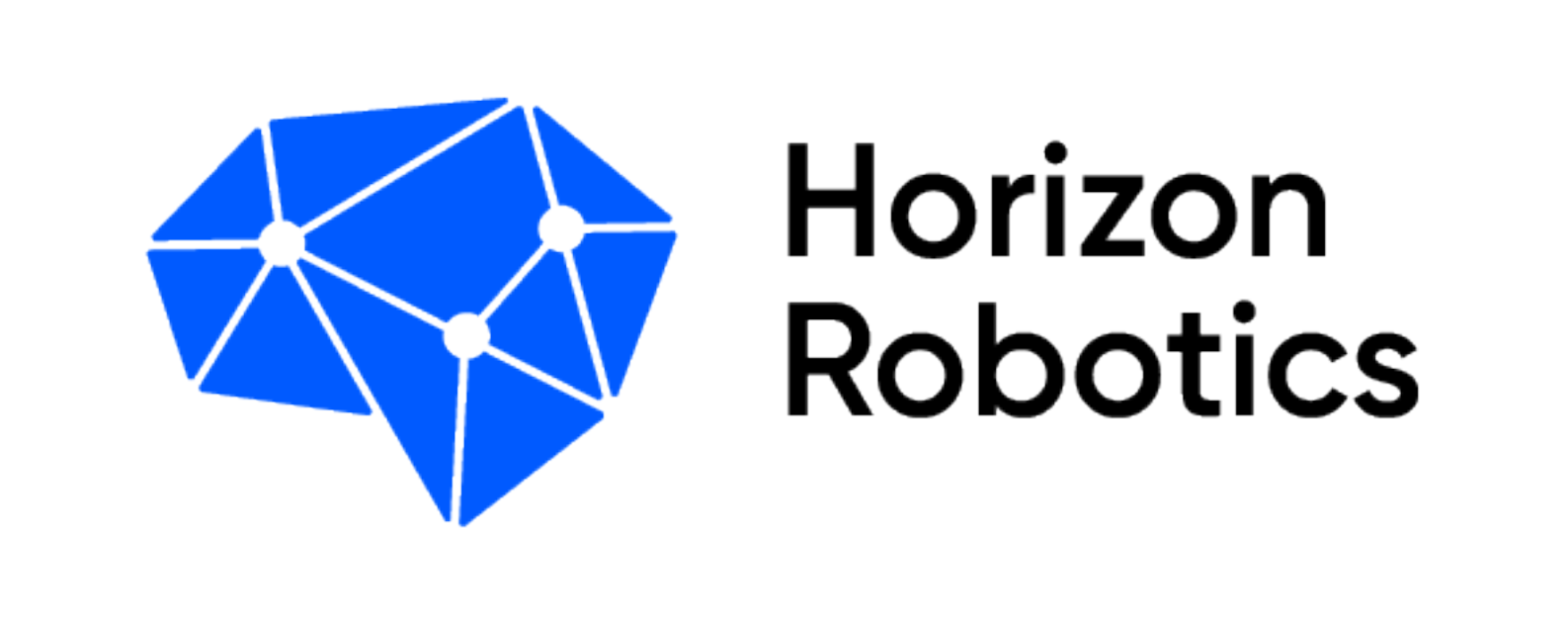}
            \hspace{-0.5em}
            \raisebox{0.3em}{\color{gray!50}\rule{0.5pt}{1.2em}} \hspace{0.5em}
            \includegraphics[height=2em]{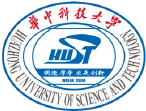}
            \hfill 
        }
        
        \par\vspace{-0.8em} 
        \noindent\textcolor{horizonblue}{\rule{\textwidth}{1pt}} 
        
        \par\vspace{0.4em}
        \centering 
        \hspace{0.15em}\raisebox{-0.35em}{\includegraphics[height=1.2\baselineskip]{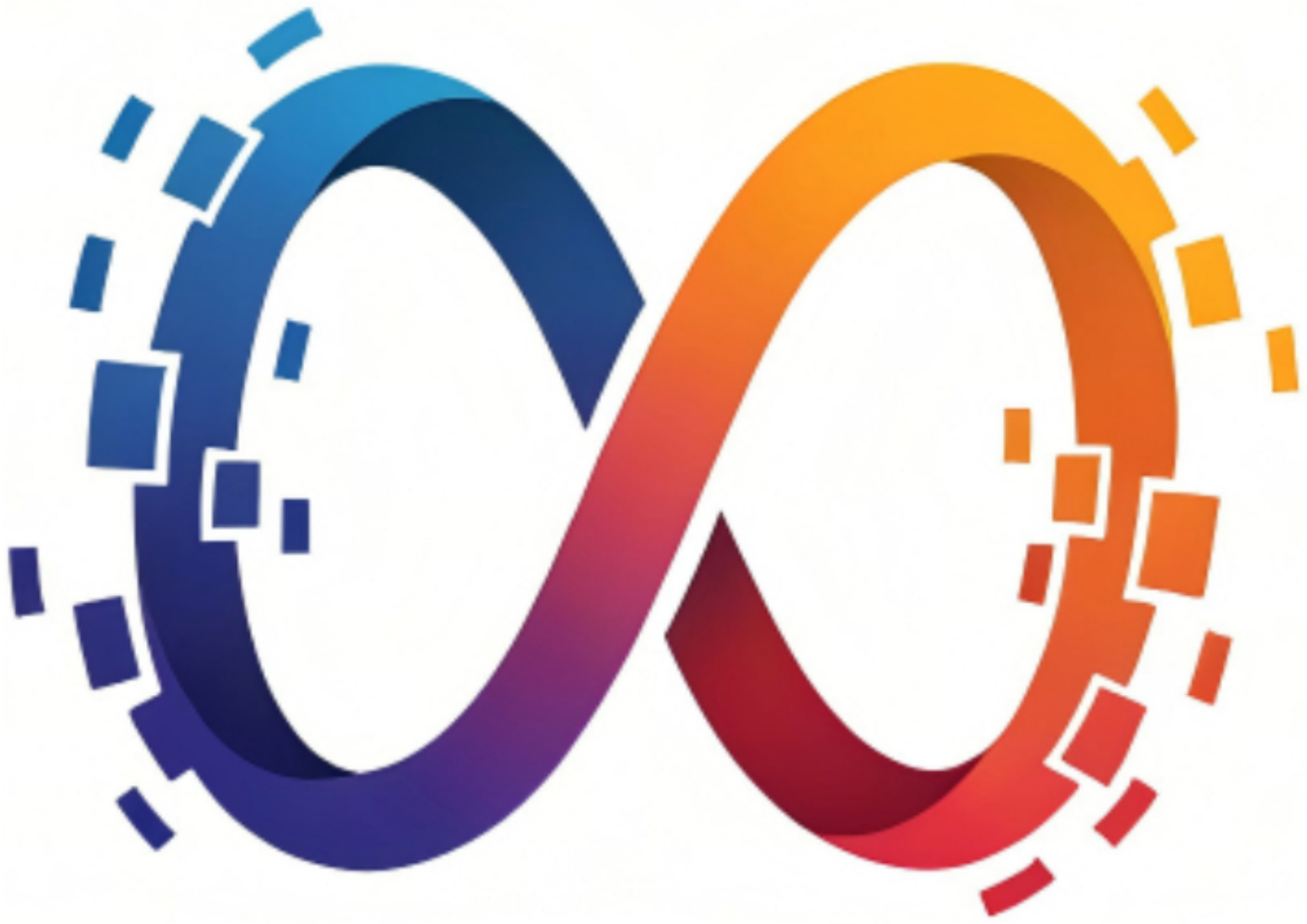}} \thename: Synergizing Linear and Sparse Attention for Highly-Efficient, Unlimited-Input Vision-Language Models
        
        \par\vspace{0.0em}
        \noindent\textcolor{horizonblue}{\rule{\textwidth}{1pt}} 
        \par\mytitlespace
    }
}

\author{
    Hongyuan Tao$^{1,\diamond}$ \quad
    Bencheng Liao$^{1}$ \quad
    Shaoyu Chen$^{2}$ \quad
    Haoran Yin$^{2}$ \quad \\ 
    Qian Zhang$^{2}$ \quad
    Wenyu Liu$^{1}$ \quad
    Xinggang Wang$^{1,\textrm{\Letter}}$ \quad \\
    $^{1}$Huazhong University of Science and Technology \quad
    $^{2}$Horizon Robotics \quad \\
    \textbf{Code \& Model \& Demo:} \href{https://github.com/hustvl/InfiniteVL}{hustvl/InfiniteVL}
}

\begin{document}

\newcommand{\mytitlespace}{\vspace{-1em}} 
\maketitle 

\renewcommand{\mytitlespace}{} 

\vspace{-2pt}

\begin{abstract}
Vision-Language Models (VLMs) are increasingly tasked with ultra-long multimodal understanding. While linear architectures offer constant computation and memory footprints, they often struggle with high-frequency visual perception compared to standard Transformers. To bridge this gap, we introduce \textbf{InfiniteVL}. We first develop a hybrid base model called \textbf{InfiniteVL-Base} that interleaves a small fraction of Full Attention layers with Gated DeltaNet. Empowered by a tailored distillation and fine-tuning strategy, InfiniteVL-Base matches the fundamental multimodal performance of equivalent Transformers while achieving a \textbf{1.7$\times$} decoding speedup. However, the quadratic complexity of the retained Full Attention inevitably becomes an efficiency bottleneck when scaling to ultra long context. To break this barrier, we propose a novel Long-Sequence Architectural Fine-Tuning strategy that seamlessly transforms the dense attention into vision-specific sparse mechanisms. This yields two specialized variants: \textbf{InfiniteVL-Offline} for offline retrieval and \textbf{InfiniteVL-Online} for online streaming. By eliminating the computation explosion of global attention without sacrificing high-frequency visual recall, InfiniteVL-Offline achieves Transformer-level length generalization with a \textbf{5x} prefill acceleration at 256K context. Concurrently, InfiniteVL-Online delivers robust streaming perception with a constant memory footprint and a real-time throughput of \textbf{25} FPS.
\end{abstract}   

\vspace{-22pt}
\begin{center}
    \captionsetup{type=figure}
    \includegraphics[width=1\textwidth]{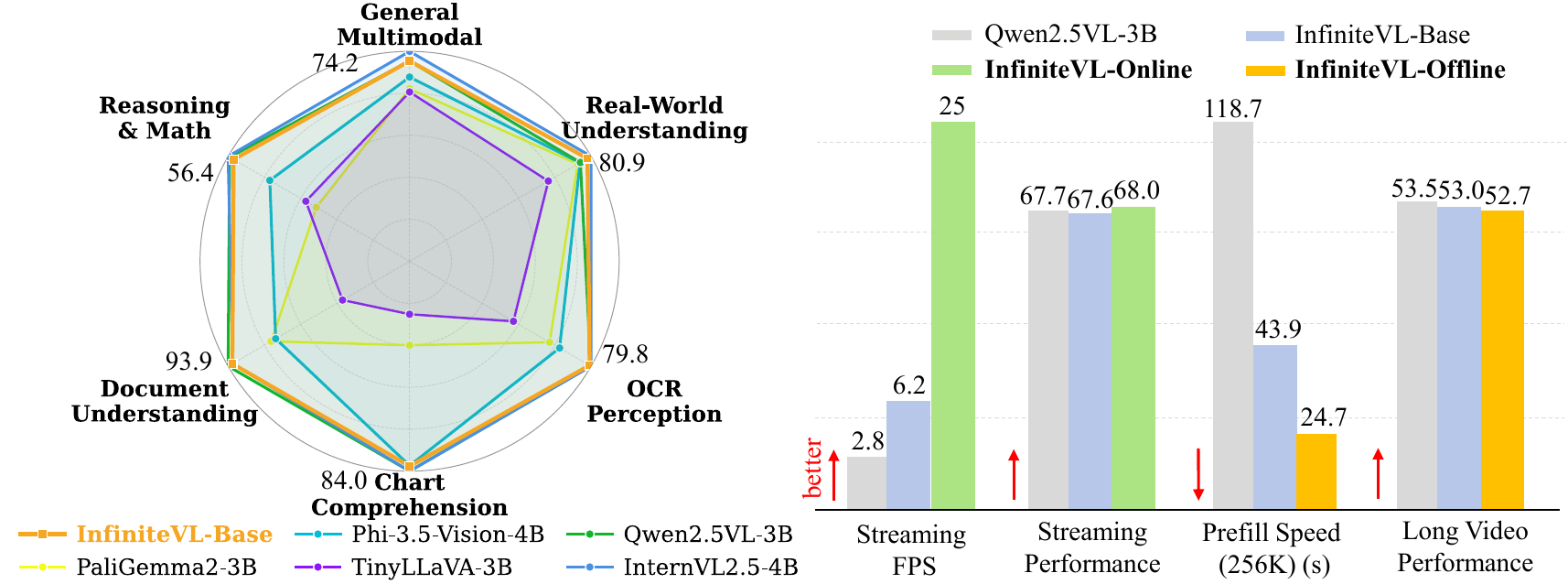}
    \vspace{-20pt}
    \captionof{figure}{
        \textbf{Performance and Efficiency Overview.} (Left) InfiniteVL-Base fully matches the foundational multimodal performance of leading 2B--4B Transformer VLMs. (Right) In ultra-long scenarios, our variants achieve a superior balance between efficiency and long-context understanding.
        }
    \label{fig: teaser}
\end{center} 
\vspace{-5pt}

\begingroup
\makeatletter
\renewcommand\@makefnmark{}
\renewcommand\thefootnote{} 
\makeatother

\footnote{
    $^\diamond$ Intern of Horizon Robotics; 
    $^\textrm{\Letter}$ Corresponding author: Xinggang Wang 
    (\textcolor{blue}{\tt\small xgwang@hust.edu.cn}).
}

\addtocounter{footnote}{-1}
\endgroup

\section{Introduction}

VLMs~\cite{liu2023visual,li2023blip,zhuminigpt,bai2025qwen2,chen2024expanding,guo2025seed1,lin2024vila, zeng2025diffusionvl} are rapidly evolving from static image-text comprehension toward continuous, real-world multimodal understanding~\cite{black2410pi0,driess2023palm,zitkovich2023rt,jiang2024senna,li2025recogdrive}. However, scaling these models to handle long multimodal inputs is severely challenged by the inherent computational constraints of the standard Transformer architecture~\cite{vaswani2017attention}. The quadratic complexity of self-attention with respect to sequence length, coupled with a dynamically growing key-value (KV) cache during autoregressive inference, creates unsustainable computational and memory demands. This inefficiency becomes a critical bottleneck in long-context scenarios, such as extended video understanding or continuous agent interaction. It also prohibits deployment on resource-constrained edge devices, where real-time performance and a strictly low memory footprint are paramount.

Recently, linear attention mechanisms~\cite{katharopoulos2020transformers, schlag2021linear, gu2024mamba, yanggla, yang2024parallelizing, yanggated, hou2025visualrwkv, dig, vig} have emerged as a promising avenue to address the long-context efficiency bottlenecks in VLMs. Benefiting from constant-size state compression and recent advances in hardware-aware efficient computation, linear architectures can process unbounded multimodal sequences with a strictly constant computation and memory footprint, circumventing the quadratic scaling of standard Transformers. However, this inherent state compression comes at a cost: it inevitably discards high-frequency visual details. Consequently, existing linear VLMs~\cite{zhao2025cobra,liao2025multimodal, li2024videomamba, li2025matvlm, qiao2024vl, zou2025omnimamba} suffer severe performance degradation on information-dense multimodal tasks, such as Optical Character Recognition (OCR) and high-resolution document comprehension, where localized precise perception is indispensable. As a result, current solutions are caught in a dilemma between long-term computational efficiency and precise visual performance.

Hybrid architectures which interleave a small fraction of Full Attention layers with linear modules have shown promise in Large Language Models (LLMs) by restoring precise retrieval capabilities~\cite{lieber2024jamba, ren2024samba, blakeman2025nemotron}. We hypothesize that extending this hybrid paradigm to multimodal domains could effectively recover the high-frequency visual details typically discarded by linear VLMs. Building on this insight, we first construct a foundational hybrid base for InfiniteVL which is called \textbf{InfiniteVL-Base}. This architecture allocates approximately 75\% of the Gated DeltaNet layers~\cite{yanggated} for efficient long-term memory compression, while retaining 25\% as Full Attention layers to preserve precise visual perception. To effectively optimize this hardware-friendly architecture under limited computational resources, we implement a tailored training strategy utilizing a carefully curated multimodal corpus. This involves \textbf{Distillation Pretraining} to transfer robust multimodal knowledge from advanced Transformers, followed by \textbf{Continuous Supervised Fine-tuning (Continuous SFT)} to enhance component alignment and instruction following. As a preliminary exploration, InfiniteVL-Base successfully achieves precise multimodal understanding on par with mainstream Transformers, delivering over a 1.7$\times$ inference speedup and a 3.1$\times$ expansion in context length processing capabilities. 

However, the quadratic memory growth of the retained Full Attention layers inevitably triggers severe bottlenecks when scaling to ultra-long multimodal sequences. To overcome this, we propose a novel \textbf{Long-Sequence Architectural Fine-Tuning strategy} that adaptively converts the dense attention into vision-specific sparse mechanisms, yielding two variants tailored for the two primary scenarios of multimodal long-sequence understanding: online streaming and offline comprehension. For online streaming, the model faces an unbounded influx of frames, strict constant-memory constraints, and an inherent recency bias. We introduce \textbf{InfiniteVL-Online}, which integrates a sliding window to capture recent high-frequency details and attention sinks~\cite{xiao2023efficient} as anchors to stabilize computation. Combined with the intact DeltaNet layers, it achieves Transformer-level performance on StreamingBench~\cite{lin2024streamingbench} while sustaining a robust, real-time throughput of 25 FPS.  Alternatively, offline comprehension ingests all frames simultaneously, incurring exorbitant quadratic prefill costs and demanding exact retrieval. Here, we propose \textbf{InfiniteVL-Offline} which utilizes dynamic top-$k$ routing mechanism that performs chunk-level retrieval, mitigating the prefill burden while preserving precise visual recall. Evaluations on Video-MME~\cite{fu2025video} and LongVideoBench~\cite{wu2024longvideobench} show that InfiniteVL-Offline matches the length generalization of mainstream Transformers, delivering comparable performance alongside a 5$\times$ prefill acceleration at a 256K context length.

In summary, our main contributions are as follows:  

\begin{itemize}     
 \item We propose \textbf{InfiniteVL}, systematically integrating Gated DeltaNet with Full Attention to bridge the precise visual perception gap of linear VLMs. To fundamentally break the quadratic efficiency bottleneck in extended contexts, we introduce a Long-Sequence Architectural Fine-Tuning strategy, adaptively converting the retained attention into vision-specific sparse variants (InfiniteVL-Online and InfiniteVL-Offline) tailored for online and offline scenarios.          
\item We design a highly effective, tailored training pipeline encompassing distillation pretraining, Continuous SFT, and Long-SFT. This strategy successfully overcomes the inherent training instability of linear layers, seamlessly transferring robust multimodal knowledge from mainstream Transformers.          
\item Extensive experiments demonstrate that InfiniteVL-Base achieves Transformer-level fundamental multimodal understanding with over a 1.7$\times$ decoding speedup. In ultra-long multimodal scenarios, InfiniteVL-Offline delivers a 5$\times$ prefill acceleration at a 256K context length without sacrificing retrieval accuracy. Concurrently, InfiniteVL-Online sustains a robust real-time throughput of 25 FPS with a strictly constant $\mathcal{O}(1)$ memory footprint. 
\end{itemize}
\section{Related Work}

\subsection{Vision-Language Models}
Modern VLMs typically integrate visual encoders with pre-trained LLMs~\cite{brown2020language,touvron2023llama} through large-scale image-text pretraining. By scaling up data and model capacity, this paradigm achieves strong performance on fundamental vision-language tasks, such as Visual Question Answering (VQA), image captioning, and visual grounding~\cite{black2410pi0,driess2023palm,zitkovich2023rt,jiang2024senna,li2025recogdrive}. Additionally, many contemporary VLMs extend their capabilities to video understanding by incorporating temporal modeling mechanisms, allowing them to process short to medium length video clips~\cite{maaz2024video,song2024moviechat,bai2025qwen2,chen2024expanding,chenlongvila,song2025videonsa}. As the field advances, a key direction for VLMs is continuous and long-term multimodal understanding particularly for embodied AI and physical scene perception~\cite{yang2025cambrian}. This requires models to process extensive spatiotemporal contexts. However, using the standard Transformer architecture for such long sequences is severely constrained by its quadratic computational complexity and the continuous growth of the KV cache during inference. While recent methods attempt to extend context length through cache dropping or sliding windows, these approaches often compromise long-term memory and situational coherence~\cite{zhang2023h2o, chen2024image}. Therefore, it is necessary to explore more memory and computation efficient architectures.

\subsection{Linear Attention}
Linear attention~\cite{katharopoulos2020transformers} and State Space Models (SSMs)~\cite{gu2024mamba} have emerged as pivotal architectures to address the quadratic complexity of standard softmax attention. By decomposing similarity computation and utilizing recurrent state updates, these methods compress historical information into a fixed size state, reducing computational complexity to linear scale while enabling inference with constant memory. To mitigate the capacity limits and memory collisions inherent in vanilla linear attention, recent advances have introduced forgetting mechanisms and delta rule-based updates. This evolution has yielded highly efficient architectures, such as Mamba~\cite{gu2024mamba}, Gated Linear Attention (GLA)~\cite{yanggla}, and Gated DeltaNet~\cite{yanggated}, which effectively enhance memory compaction and serve as robust repositories for long-term context.

Motivated by these computational benefits, several prior works have explored VLMs build with linear attention or SSMs~\cite{zhao2025cobra,liao2025multimodal,li2025matvlm,li2024videomamba}. However, applying those architectures to multimodal tasks exposes a critical limitation. The inherent state compression mechanism inevitably causes the loss of high-frequency visual details. Consequently, existing linear VLMs often underperform on information-dense perception tasks, such as OCR and high-resolution document understanding when compared to Transformer baselines. This performance gap indicates that relying solely on compressed recurrent states is insufficient for comprehensive visual perception, highlighting the necessity for architectures that can balance efficient memory compression with precise local detail retention.

\subsection{Hybrid Architecture}
The weakness of linear attention models in exact recall tasks inherently stems from their low-rank state compression nature. While effectively retaining macroscopic long-term memory, they inevitably discard high-frequency visual details. To mitigate this, some works have explored hybrid architectures by reintroducing a small fraction of Full Attention layers, specifically tasked with recovering the precise features lost during state compression~\cite{lieber2024jamba, ren2024samba, glorioso2024zamba}. Empirical studies in LLMs have demonstrated that interleaving even a minimal proportion of Full Attention can effectively bridge the retrieval capability gap of linear architectures~\cite{lieber2024jamba}, thereby driving the design and scaling of large-scale hybrid LLMs~\cite{blakeman2025nemotron, chen2025minimax, team2025kimi, team2026minicpm}. 

Recently, this hybrid paradigm has been extended to VLMs~\cite{liao2025multimodal, li2025matvlm}. However, they still exhibit a noticeable performance gap compared to leading Transformer-based VLMs. Concurrently, the reintroduced Full Attention layers reintroduces an efficiency bottleneck for multimodal long-sequence understanding. Therefore, exploring a highly efficient multimodal hybrid architecture that can simultaneously bridge the visual perception gap and demonstrate compelling efficiency advantages has become a critical open challenge.

\begin{figure*}[ht!]
    \centering
    \includegraphics[width=0.95\linewidth]{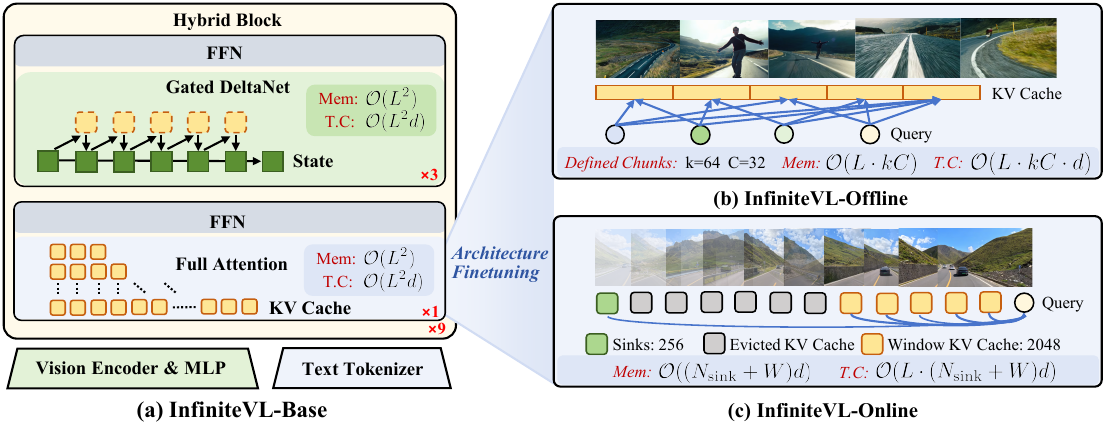}
    \caption{\textbf{Overall Architecture of InfiniteVL.} (a) InfiniteVL-Base hybridizes Gated DeltaNet with a small fraction of Full Attention. Through our Long-Sequence Architectural Fine-Tuning, we yield two highly efficient variants: (b) InfiniteVL-Offline for efficient block-level routing in offline long-contexts, and (c) InfiniteVL-Online for constant-memory online streaming.}
    \label{fig: architecture}
\end{figure*}

\section{Method}

As illustrated in \cref{fig: architecture}, InfiniteVL is a high-performance, long-context hybrid VLM. We first construct InfiniteVL-Base (\cref{sec:base}) by integrating a vision encoder with an LLM backbone that interleaves full and linear attention, optimized via Distillation Pretraining and Continuous SFT (\cref{sec:training}). For ultra-long sequences, we introduce a long-sequence architectural fine-tuning stage (\cref{sec:long_sft}). By adaptively converting global attention into vision-specific sparse mechanisms, we derive two variants: InfiniteVL-Online for online streaming and InfiniteVL-Offline for offline contexts, achieving a better balance between context efficiency and precise visual perception.

\subsection{InfiniteVL-Base}
\label{sec:base}
InfiniteVL-Base comprises a Qwen2.5-VL~\cite{bai2025qwen2} vision encoder, a projection MLP, and a decoder-only LLM backbone containing 9 hybrid blocks ($d=2048$). To balance precise retrieval and efficiency, each block interleaves one Full Attention layer with three consecutive Gated DeltaNet layers, connected via residual pathways and Pre-Layer Normalization with SiLU activations. The Full Attention layer uses Grouped-Query Attention (GQA)~\cite{ainslie2023gqa} with 16 Query and 2 KV heads. Spatial-temporal semantics are injected via 3D Rotary Position Embedding (3D RoPE)~\cite{su2024roformer}. The standard attention is computed as:

\begin{equation} 
\mathbf{O}_{\text{FA}} = \text{Softmax}\left(\frac{\mathbf{Q}\mathbf{K}^T}{\sqrt{d_k}}\right)\mathbf{V}. 
\end{equation} 

Gated DeltaNet layers\cite{yanggated} provide constant-memory state compression. Since Gated DeltaNet intrinsically avoids KV cache explosion, we can afford to symmetrically configure it with 16 Query and 16 KV heads, thereby maximizing its outer-product memory capacity. We further integrate a 1D convolution (window size 4) and an output gate for expressiveness. Operating without explicit positional encodings or weight biases, the recurrent memory update and retrieval are formulated as:

\begin{align}
 \mathbf{S}_t =\gamma_t  &\mathbf{S}_{t-1} + \beta_t \mathbf{K}_t \otimes (\mathbf{V}_t - \mathbf{S}_{t-1} \mathbf{K}_t),\\
& \mathbf{O}_{\text{Linear}, t} = \mathbf{Q}_t \mathbf{S}_t,
\end{align}
where $\mathbf{S}_t$ is the fixed-size hidden state, $\gamma_t$ serves as a data-dependent decay gate, and $\beta_t$ controls the update rate.

\begin{figure*}[ht!]
    \centering
    \includegraphics[width=1.0\linewidth]{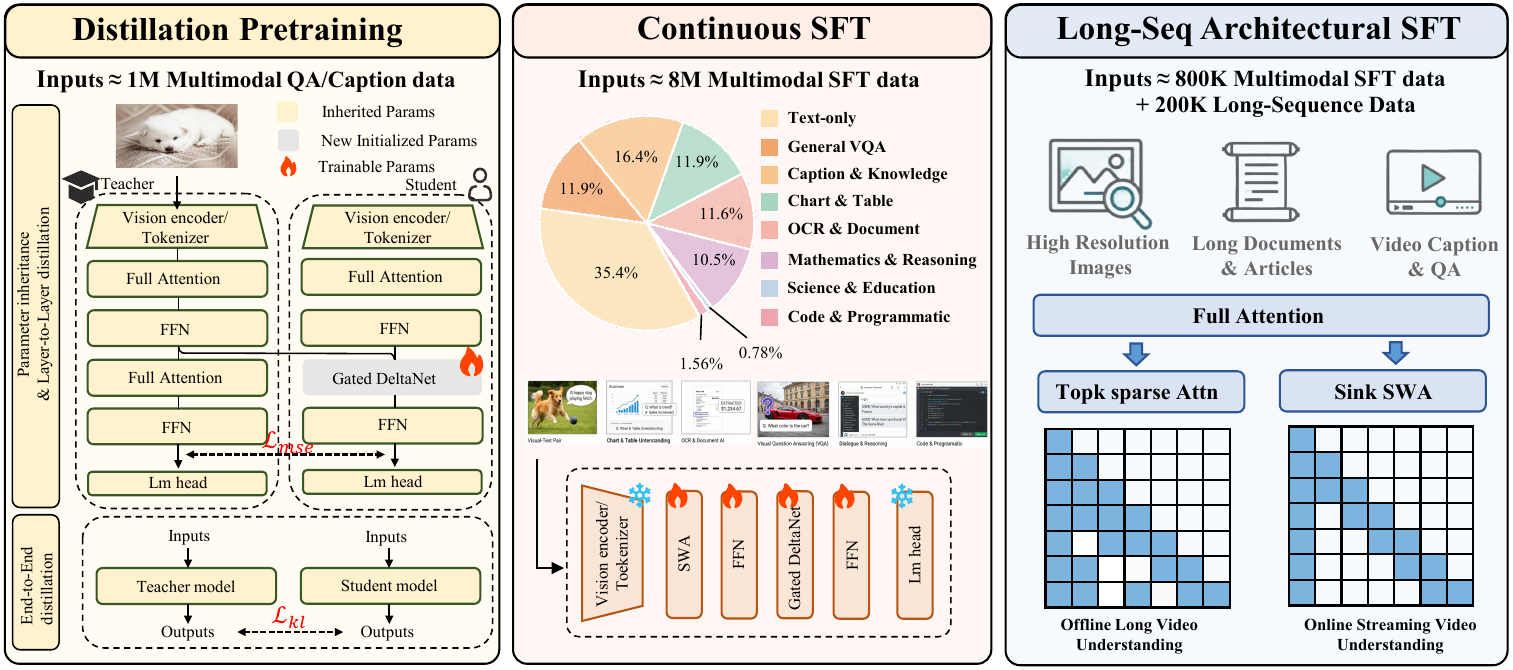}
    \caption{\textbf{Progressive Multi-Stage Training Strategy.} Stage I employs layer-to-layer and end-to-end distillation to align the newly initialized linear modules; Stage II enhances component alignment via Continuous SFT; Stage III specifically adapts the dense attention into vision-specific sparse mechanisms for extended contexts.}
    \label{fig: train}
\end{figure*}

\subsection{Efficient Multi-Stage Training Strategy} 
\label{sec:training}  
Training a hybrid VLM from scratch is computationally prohibitive and demands massive multimodal corpora. To circumvent this, we aim to leverage the robust pre-trained knowledge in leading Transformer VLMs. Consequently, we adopt a minimal-replacement strategy, preserving the majority of the pre-trained weights while selectively substituting specific dense attention layers with randomly initialized linear modules. However, this structural modification introduces severe architectural heterogeneity, rendering direct optimization highly unstable and prone to catastrophic forgetting. To efficiently align these components and stabilize the training process, we design a progressive two-stage training pipeline for InfiniteVL-Base, as illustrated in \cref{fig: train}. By decoupling the learning objectives, this paradigm first employs Distillation Pretraining to stabilize the newly initialized linear modules, followed by Continuous SFT to achieve seamless component alignment and unlock complex multimodal interactions. This phased approach guarantees the seamless transfer of robust pre-trained priors while strictly constraining the overall training budget.

\subsubsection{Training Data.} 
We curate a comprehensive multimodal corpus integrating \textit{FineVision}~\cite{wiedmann2025finevision}, \textit{LLaVA-OneVision-1.5}~\cite{an2025llava}, \textit{PixMo}~\cite{deitke2024molmo}, \textit{The Cauldron}~\cite{laurençon2024matters}, and \textit{Docmatix}~\cite{laurençon2024building}. Spanning tasks from OCR to mathematics, its scale and diversity match leading VLMs like InternVL2.5~\cite{chen2024expanding}.

\subsubsection{Distillation Pre-training.} 
We initialize the student by replacing specific Qwen2.5-VL attention layers with Gated DeltaNet. To align these modules, we first perform layerwise feature distillation, minimizing the Mean Squared Error (MSE) between the teacher's ($h^{(i)}_{\text{Trans}}$) and student's ($h^{(i)}_{\text{GDN}}$) $i$-th layer outputs given the same input:
\begin{equation}
\mathcal{L}^{(i)}_{\text{layer}} = \bigl\lVert h^{(i)}_{\text{GDN}} - h^{(i)}_{\text{Trans}} \bigr\rVert_2^{2}. 
\end{equation}
This ``same-input, output-matching'' design accelerates alignment. Next, an end-to-end distillation phase minimizes the Kullback-Leibler (KL) divergence between token-level logits:
\begin{equation}
\mathcal{L}_{\text{logit}} = \frac{1}{T}\sum_{t=1}^{T} \mathrm{KL}\!\left(\operatorname{Softmax}(z^{T}_{t}) \,\big\|\, \operatorname{Softmax}(z^{S}_{t})\right), 
\end{equation}
where $z^{T}_{t}$ and $z^{S}_{t}$ are the teacher and student logits. Here, we cap image resolution at $512 \times 512$ and input length to 8192 tokens.

\subsubsection{Continuous SFT.} 
To unlock the linear modules' full potential and mitigate exposure bias~\cite{pozzi2025mitigating}, we fine-tune the distilled model using a standard Cross-Entropy (CE) loss:
\begin{equation}
\mathcal{L}_{\mathrm{SFT}} = \frac{1}{T} \sum_{t=1}^{T} \mathrm{CE}\big(q_t, \operatorname{Softmax}(z_t^{S})\big), 
\end{equation}
where $q_t$ is the target distribution. To improve precise visual detail perception, we increase maximum resolution to $1344 \times 1344$ while maintaining the 8192-token limit.

\subsection{Long-Sequence Architectural Fine-Tuning Strategy} 
\label{sec:long_sft} 

By interleaving a small fraction of Full Attention layers, InfiniteVL-Base effectively mitigates the detail forgetting issues inherent in linear architectures. However, when scaling to ultra-long multimodal sequences of length $L$, the retained Full Attention layers inevitably dominate the computational and memory footprint, scaling quadratically $\mathcal{O}(L^2)$ in prefill time and linearly $\mathcal{O}(L)$ in decoding memory. We identify that realworld long-context multimodal understanding predominantly unfolds in two distinct paradigms: \textit{offline comprehension} and \textit{online streaming}, each presenting unique efficiency bottlenecks. To overcome these without compromising the model's precise visual perception or causing catastrophic forgetting of pre-trained knowledge, we propose a Stage III Long-Sequence Architectural Fine-Tuning strategy. Tailored to the specific characteristics of the two paradigms, we introduce two variants: InfiniteVL-Offline and InfiniteVL-Online. These variants adaptively transform the dense global attention into vision-specific sparse mechanisms, acting as highly efficient retrieval complements to the linear states while fundamentally shifting the complexity paradigm.

\subsubsection{InfiniteVL-Offline for Offline Comprehension.}  
In offline comprehension tasks (e.g., long-video analysis), all multimodal tokens are ingested simultaneously. This upfront loading triggers the $\mathcal{O}(L^2 d)$ quadratic prefill complexity of the retained Full Attention, leading to a severe computational cost that dominates overall latency. Furthermore, these tasks often require exact retrieval across the massive context (e.g., pinpointing a specific visual event).

To address this severe prefill bottleneck while satisfying the demand for exact retrieval, we introduce InfiniteVL-Offline, which reuses dense attention parameters through a parameter-free architecture modification. Inspired by InfLLM-V2~\cite{zhao2025infllm}, we implement a coarse-to-fine compression paradigm to perform efficient block-level routing. Specifically, the historical sequence is partitioned into fixed-size chunks of 32 tokens. For a given query block $i$, we first compute coarse-grained attention scores and dynamically route queries to the top-$k$ ($k=64$) most relevant blocks. The final attended indices are mathematically defined as the union of three critical components:

\begin{equation} 
\mathcal{I}(i) = \mathcal{I}_{\text{init}} \cup \mathcal{I}_{\text{local}}(i) \cup \mathcal{I}_{\text{topk}}(i). 
\end{equation} 
Here, $\mathcal{I}_{\text{init}}$ protects the initial system prompts, $\mathcal{I}_{\text{local}}(i)$ preserves local temporal coherence, and $\mathcal{I}_{\text{topk}}(i)$ enables global precise retrieval. By restricting the attention computation to only these retrieved blocks (totaling length $C \ll L$), the prefill time complexity is drastically reduced from $\mathcal{O}(L^2 d)$ to $\mathcal{O}(L \cdot C d)$, effectively achieving linear scaling. This hybrid dense-sparse design relies on the linear layers to compress the macroscopic long-term history, while the sparse attention efficiently routes and recalls specific high-frequency visual details. 

\subsubsection{InfiniteVL-Online for Online Streaming.}  
Conversely, online streaming (e.g., continuous agent perception) is characterized by an unbounded, continuous influx of frames. In this paradigm, standard attention requires $\mathcal{O}(L d)$ memory for the KV cache, which eventually exceeds GPU VRAM capacity and inevitably leads to Out-of-Memory (OOM) failures. Therefore, the memory footprint must be strictly bounded to a constant $\mathcal{O}(1)$ scale.

While reducing Full Attention to standard Sliding Window Attention (SWA) successfully bounds memory and aligns well with the inherent \textit{recency bias} of streaming perception, naive SWA suffers from "attention collapse". It indiscriminately discards early historical tokens (e.g., system instructions and initial scene setups) which absorb massive attention scores and are paramount for maintaining the stable distribution of the softmax function~\cite{xiaoefficient, xu2025streamingvlm}.

To achieve continuous streaming without performance degradation, InfiniteVL-Online utilizes a sliding window to capture recent high-frequency details while designating a fixed length of initial tokens as ``attention sinks'' serving as anchors to stabilize the computation. The attention mask matrix $M$ is formulated as:

\begin{equation}  
M_{i,j} =   \begin{cases}   1, & \text{if } j < N_{\text{sink}} \text{ or } i - j < W, \\   0, & \text{otherwise},   \end{cases}  
\end{equation}  
where $N_{\text{sink}} = 256$ is the number of sink tokens and $W = 2048$ is the window size. Because the retained KV cache size is strictly bounded to $N_{\text{sink}} + W$, the memory complexity drops from $\mathcal{O}(L d)$ to $\mathcal{O}(1)$, and the decoding time per token becomes constant. Combined with the intact Gated DeltaNet layers that continuously compress distant history into a fixed-size state, InfiniteVL-Online guarantees sustainable processing, completely eliminating OOM risks.

\subsubsection{Long-Sequence Fine-Tuning.} 

Both variants seamlessly inherit the pre-trained weights of InfiniteVL-Base, optimizing long-sequence efficiency without harming foundational multimodal comprehension. To further enhance length generalization up to 32,768 tokens, we conduct a third stage of fine-tuning as shown in \cref{fig: train}. The training data blends 800,000 samples from the Stage II SFT corpus with 200,000 video QA pairs uniformly sampled from LLaVA-Video-178K~\cite{zhang2024llava} (10 FPS, up to 224 frames, max 256 tokens/frame). The optimization objective strictly mirrors the Cross-Entropy loss of the SFT stage, smoothly adapting the model to the newly introduced sparse attention masks.


\begin{table*}[t]
\centering

\begin{subtable}{\textwidth}
\centering
\resizebox{\textwidth}{!}{%
\begin{tabular}{l|ccccccc|c}
\toprule
\textbf{Model}
& \textbf{MME}
& \textbf{MMStar}
& \textbf{MMBench\textsubscript{test en}}
& \textbf{SeedBench\textsubscript{image}}
& \textbf{ScienceQA\textsubscript{val}}
& \textbf{RealworldQA}
& \textbf{AI2D\textsubscript{w/o M}} 
& \textbf{Average} \\

\midrule

TinyLLaVA-3B~\cite{zhou2024tinyllava}    & 1733 & 37.9 & 69.5 & 70.2 & 68.7 & 55.0 & 61.8 & 61.8 \\
PaliGemma2-3B~\cite{google_pali_gemma2_modelcard_2025}    & 1658 & 52.7 & 60.7 & 71.6 & \underline{94.3} & 58.3 & 72.2 & 68.0 \\
Phi-3.5-Vision-4B~\cite{microsoft_phi35vision_hf_2025} & 1846 & 47.5 & 76.0 & 71.2 & 92.2 & 57.9 & 77.8 & 70.9 \\
SmolVLM2-2B~\cite{marafioti2025smolvlm}      & 1764 & 46.0 & 43.0 & 70.9 & 90.0 & 58.4 & 74.9  & 64.8 \\
InternVL2.5-4B~\cite{chen2024expanding}  & \textbf{2338} & \textbf{58.3} & \textbf{81.1} & \textbf{74.1} & \textbf{97.0} & 64.3 & \underline{81.4} & \textbf{78.5} \\
Qwen2.5VL-3B~\cite{bai2025qwen2}    & \underline{2171} & 54.3 & 78.2 & \underline{73.3} & 81.4 & 65.4 & \textbf{81.6} & 74.4 \\
Qwen2.5VL*(4B) & 2089 & \underline{55.7} & 78.7 & 72.9 & 93.1 & \underline{65.8} & 77.1 & 75.3 \\
\rowcolor{blue!10}
InfiniteVL-Base  & 2126 & 55.6 & \underline{79.0} & 72.9 & 93.4 & \textbf{67.3} & 77.2 & \underline{75.8} \\
\bottomrule
\end{tabular}}
\caption{Results on General Benchmarks} 
\label{tab:sub_results1}
\end{subtable}


\begin{subtable}{\textwidth}
\centering
\resizebox{\textwidth}{!}{%
\begin{tabular}{l|cccccc|c}
\toprule
\textbf{Model}
& \textbf{ChartQA\textsubscript{test}}
& \textbf{TextVQA\textsubscript{val}}
& \textbf{DocVQA\textsubscript{test}}
& \textbf{OCRBench}
& \textbf{MMMU\textsubscript{val}}
& \textbf{MathVista\textsubscript{mini}} 
& \textbf{Average} \\

\midrule

TinyLLaVA-3B~\cite{zhou2024tinyllava}    & 21.2 & 55.3 & 34.7 & 36.0 & 36.2 & 28.3 & 35.3 \\
PaliGemma2-3B~\cite{google_pali_gemma2_modelcard_2025}    & 33.6 & 63.0 & 71.6 & 60.1 & 30.3 & 27.7 & 47.7 \\
Phi-3.5-Vision-4B~\cite{microsoft_phi35vision_hf_2025} & 81.8 & 72.0 & 69.3 & 59.9 & 43.0 & 43.9 & 61.7 \\
SmolVLM2-2B~\cite{marafioti2025smolvlm}      & 68.8 & 73.2 & 80.0 & 72.9  & 42.0 & 51.5 & 64.7 \\
InternVL2.5-4B~\cite{chen2024expanding}  & \textbf{84.0} & 76.8 & 91.6 & \textbf{82.8} & \textbf{52.3} & 60.5 & \underline{74.7} \\
Qwen2.5VL-3B~\cite{bai2025qwen2}    & \textbf{84.0} & \textbf{79.6} & \textbf{93.9} & 79.7 & \underline{49.6} & 62.3 & \textbf{74.9} \\
Qwen2.5VL*(4B) & \underline{82.5} & \underline{79.1} & \underline{92.2} & \underline{80.5} & 43.7 & \underline{65.2} & 73.8  \\
\rowcolor{blue!10}
InfiniteVL-Base  & 82.0 & 78.5 & 91.7 & 79.8 & 44.0 & \textbf{65.4} & 73.6 \\
\bottomrule
\end{tabular}}
\caption{Results on Chart, Document, and Math Benchmarks}
\label{tab:sub_results2}
\end{subtable}

\vspace{-0.2cm}

\caption{\textbf{Foundational Multimodal Performance.} Comparison of InfiniteVL-Base against representative 2B--4B VLMs across (a) general multimodal benchmarks and (b) visually-intensive tasks requiring precise recall (e.g., OCR, Chart, Document). InfiniteVL successfully bridges the performance gap inherent in linear architectures.} 
\label{tab:combined_results}
\vspace{-0.5cm}
\end{table*}

\section{Experiments} 
\label{sec:experiments}  


\subsection{Experimental Setup}
\label{sec:exp_setup}

\subsubsection{Baselines.}
To isolate architectural effects, we construct a controlled Transformer baseline. Using identical training pipelines and data, we train a pure Transformer VLM (initialized from Qwen2.5-VL) equivalent in scale to our model, denoted as Qwen2.5-VL*. To contextualize our performance, we also compare InfiniteVL-Base against leading 2B--4B VLMs: TinyLLaVA-3B~\cite{zhou2024tinyllava}, PaliGemma2-3B~\cite{google_pali_gemma2_modelcard_2025}, Phi-3.5-Vision-4B~\cite{microsoft_phi35vision_hf_2025}, SmolVLM2-2B~\cite{marafioti2025smolvlm}, InternVL2.5-4B~\cite{chen2024expanding}, and the official Qwen2.5-VL-3B~\cite{bai2025qwen2}. For long-context tasks, we apply an identical long-sequence fine-tuning recipe to both Qwen2.5-VL* and InfiniteVL-Base to establish rigorous comparative baselines.

\begin{table*}[t]
\centering
\small
\setlength{\tabcolsep}{3.5pt}
\renewcommand{\arraystretch}{0.95}
\resizebox{\textwidth}{!}{%
\begin{tabular}{c|ccc|cc|c}
\toprule
\multirow{2}{*}{\textbf{Model}} &
\multicolumn{3}{c|}{\textbf{High Resolution}} &
\multicolumn{2}{c|}{\textbf{Long Video}} &
\multicolumn{1}{c}{\textbf{Streaming Video}} \\
\cmidrule(lr){2-4}\cmidrule(lr){5-6}\cmidrule(lr){7-7}
&
\textbf{Vstar} &
\textbf{BLINK} &
\textbf{HRBench4K} &
\textbf{Video-MME} &
\textbf{LongVideoBench} &
\textbf{StreamingBench} \\
\midrule
Qwen2.5-VL*~ & 64.9 & 47.2 & 63.5 &  56.0 & 50.7 & 67.7 \\
InfiniteVL-Base (wo LSFT) & 62.8 & 46.4 & 53.3 & 45.3 & 42.6 & 64.9 \\
InfiniteVL-Base & 64.5 & 47.6 & 62.8 & 55.7 & 50.3 & 67.6 \\
InfiniteVL-SWA & 56.5 & 46.1 & 57.0 & 51.0 & 47.2 & 66.7 \\
InfiniteVL-Online & 58.0 & 46.5 & 59.0 & 52.1 & 48.0 & 68.0 \\
InfiniteVL-Offline & 64.1 & 47.4 & 62.5 & 55.3 & 50.1 & 67.6 \\
\bottomrule
\end{tabular}}

\vspace{0.1cm}

\caption{\textbf{Multimodal Long-Context Performance.} Comprehensive evaluation of different architectural variants on high-resolution image understanding, offline long-video comprehension, and online streaming perception benchmarks.}
\label{tab:longcontext}
\vspace{-0.5cm}
\end{table*}

\begin{figure*}[t!]
    \centering
    \includegraphics[width=1.0\linewidth]{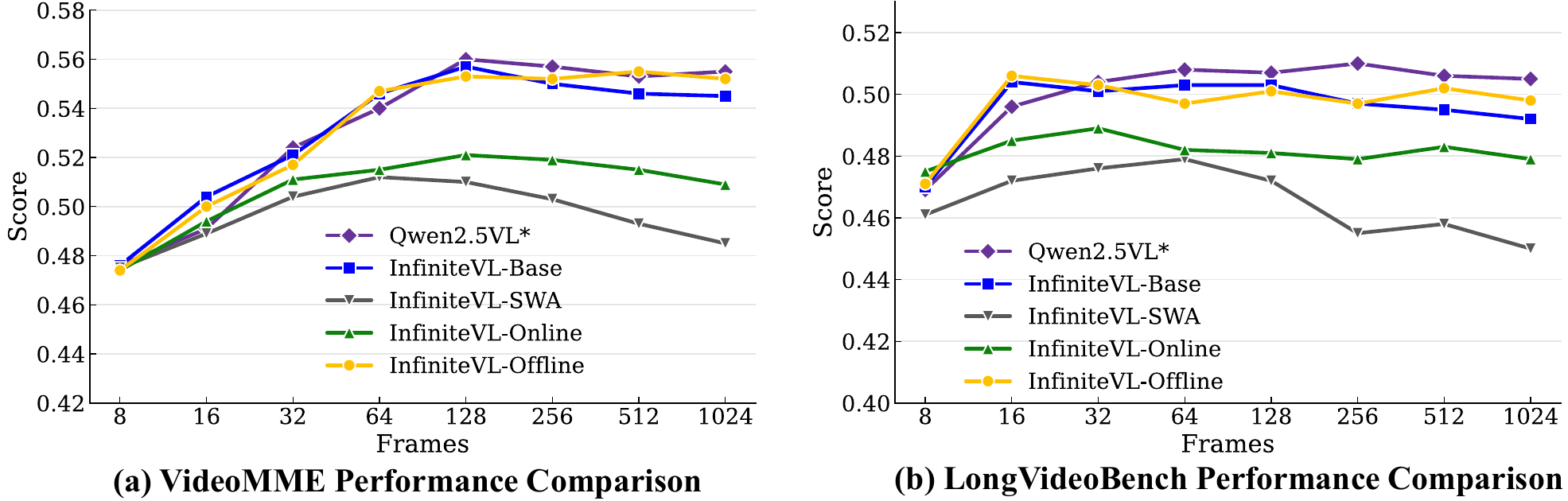}
    \caption{\textbf{Length Generalization on Long Videos.} Performance scaling across different frame counts (up to 1,024 frames) on (a) Video-MME and (b) LongVideoBench. InfiniteVL-Offline perfectly matches the robust length extrapolation trajectory of the Hybrid and Transformer baseline.}
    \label{fig: longcontext}
\end{figure*}

\subsubsection{Benchmarks.}
We assess foundational multimodal performance using VLMEvalKit~\cite{duan2024vlmevalkit} across 13 diverse public benchmarks covering general understanding, OCR, charts, documents, and reasoning: MME\cite{fu2025mme}, MMStar\cite{chen2024we}, MMBench\textsubscript{test en}\cite{liu2024mmbench}, SeedBenchIMG\cite{li2024seed}, ScienceQA\textsubscript{val}\cite{saikh2022scienceqa}, RealworldQA\cite{xai_realworldqa_2024}, AI2D\textsubscript{w/o M}\cite{kembhavi2016diagram}, TextVQA\textsubscript{val}\cite{singh2019towards}, OCRBench\cite{liu2024ocrbench}, ChartQA\textsubscript{test}\cite{masry2022chartqa}, DocVQA\textsubscript{test}\cite{mathew2021docvqa}, MMMU\textsubscript{val}\cite{yue2024mmmu} and MathVista\cite{lu2023mathvista}. For extended-context evaluation, we target three demanding scenarios: 1) \textit{High-resolution images}: Vstar~\cite{wu2024v}, BLINK~\cite{fu2024blink}, and HRBench4K~\cite{wang2025divide}; 2) \textit{Offline long-videos}: Video-MME~\cite{fu2025video} and LongVideoBench~\cite{wu2024longvideobench}; and 3) \textit{Online streaming}: StreamingBench~\cite{lin2024streamingbench}.

\subsection{Main Results of InfiniteVL-Base} 

\subsubsection{Foundational Multimodal Performance.}

As shown in \cref{tab:combined_results}(a-b), InfiniteVL-Base achieves consistent parity with our strictly controlled Transformer baseline (Qwen2.5-VL*) and leading models of similar scale. Notably, it excels in visually intensive tasks (OCR, Chart, Document understanding) that historically challenge linear VLMs. This confirms that our hybrid design successfully leverages Gated DeltaNet for macroscopic compression while retaining a fraction of Full Attention for precise detail recall, ensuring robust general-purpose perception without task-specific inductive biases.

\subsubsection{Efficiency Analysis.}
\label{base efficiency analysis}
\begin{figure*}[t!]
    \centering
    \includegraphics[width=1.0\linewidth]{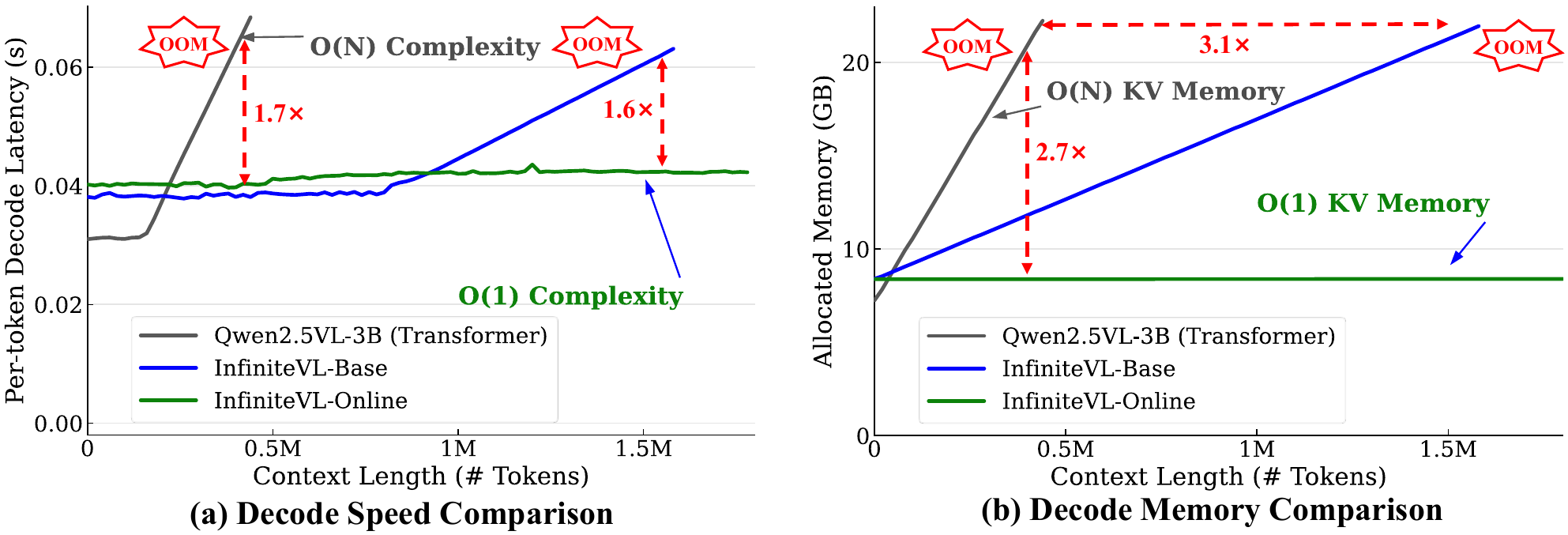}
    \caption{\textbf{Autoregressive Decoding Efficiency.} Comparison of (a) per-token decode latency and (b) allocated memory. InfiniteVL-Base successfully extends the maximum decoding length to 3.1$\times$ that of the standard Qwen2.5-VL-3B. Furthermore, empowered by its strictly constant $\mathcal{O}(1)$ memory footprint, InfiniteVL-Online unlocks theoretically infinite decoding lengths.}
    \label{fig: efficiency_1}
\end{figure*}

Evaluations are conducted on a single NVIDIA RTX 4090 with Flash Attention 2~\cite{dao2023flashattention}. \cref{fig: efficiency_1} highlights the decoding advantage: while the official Qwen2.5-VL-3B hits OOM at 454K context due to KV cache accumulation, InfiniteVL-Base scales to 1,589K tokens before OOM, owing to Gated DeltaNet's constant memory state. Furthermore, its per-token latency grows significantly slower, achieving a 1.7$\times$ speedup at 450K context. For prefilling at 256K context, InfiniteVL-Base delivers a 2.8$\times$ acceleration over Qwen2.5-VL-3B (\cref{fig: efficiency_2}(a)).

\subsection{Scaling to Extended Contexts} 
\label{sec:scaling_extended}  

\subsubsection{Multimodal Long-context Performance.} 
We compare the performance of Qwen2.5-VL*, InfiniteVL-Base, InfiniteVL-SWA, InfiniteVL-Online, and InfiniteVL-Offline on high-resolution image understanding, offline long-video comprehension, and online streaming perception tasks after identical long-context fine-tuning. Qwen2.5-VL* and InfiniteVL-Base demonstrate comparable performance across multiple benchmarks. As shown in \cref{tab:longcontext}, while InfiniteVL-SWA suffers performance degradation due to its strictly restricted attention window, InfiniteVL-Online effectively mitigates this long-context degradation. Specifically, InfiniteVL-Online matches or even exceeds the performance of InfiniteVL-Base on the StreamingBench online video task. Meanwhile, InfiniteVL-Offline maintains strict performance parity with InfiniteVL-Base across all evaluated benchmarks. 
Furthermore, \cref{fig: longcontext} provides a detailed comparison of the length generalization capabilities across these models. The evaluation is conducted on the Video-MME and LongVideoBench datasets, configured at 1 fps with 256 tokens per frame, and sampling from 8 up to 1024 frames (corresponding to context lengths ranging from 2K to 256K). The experimental results demonstrate that InfiniteVL-Online consistently exhibits significantly milder performance degradation than InfiniteVL-SWA as the sequence grows. Concurrently, InfiniteVL-Offline delivers equivalent performance to InfiniteVL-Base across all evaluated context lengths.

\begin{figure*}[t!]
    \centering
    \includegraphics[width=1.0\linewidth]{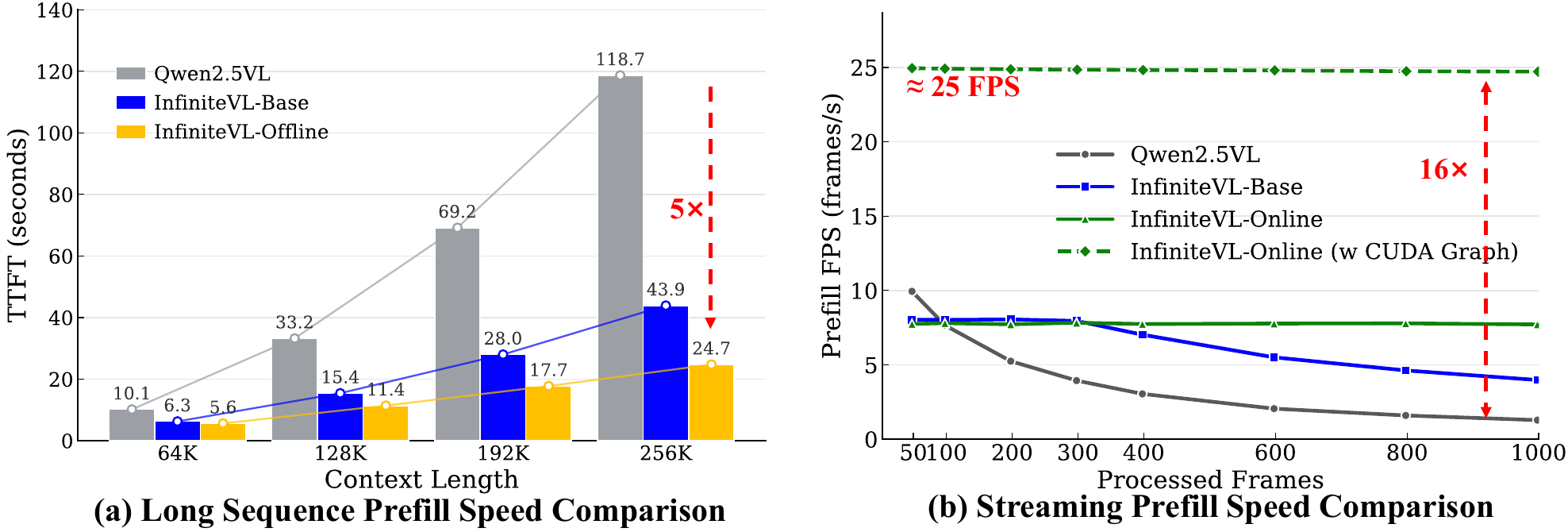}
    \caption{\textbf{Long-Sequence Prefill and Streaming Efficiency.} (a) Time-to-First-Token (TTFT) for offline prefilling, where InfiniteVL-Offline achieves a 5$\times$ acceleration at 256K context. (b) Sustained throughput (FPS) in online streaming, where InfiniteVL-Online (with CUDA Graph) maintains a robust 25 FPS without OOM degradation.}
    \label{fig: efficiency_2}
\end{figure*}

\subsubsection{Offline and Online Video Understanding Efficiency.} While \cref{base efficiency analysis} demonstrates the InfiniteVL-Base's efficiency gains over standard Transformer VLMs, InfiniteVL-Online and InfiniteVL-Offline further eliminate the computational and memory bottlenecks of the remaining Full Attention layers.   As shown in \cref{fig: efficiency_2}(b), in online streaming scenarios, InfiniteVL-Online achieves a stable inference speed of 8 FPS. Crucially, its constant memory footprint enables the use of CUDA Graph acceleration, which propels the throughput to a real-time streaming rate of 25 FPS. Conversely, InfiniteVL-Base and Qwen2.5-VL* suffer from steadily decreasing FPS and eventual OOM errors due to their unconstrained KV cache growth. \cref{fig: efficiency_2}(a) shows that InfiniteVL-Offline drastically reduces the prefill computational load for offline processing. At a 256K input length, it demonstrates a 1.8$\times$ and 5$\times$ prefill speedup over InfiniteVL-Base and Qwen2.5-VL* respectively.

\begin{table*}[t]
\centering
\large
\setlength{\tabcolsep}{3.5pt}
\renewcommand{\arraystretch}{0.95}

\newcommand{\rot}[1]{\rotatebox[origin=bl]{45}{\textbf{#1}}}

\resizebox{\textwidth}{!}{%
\begin{tabular}{l|cccc|cccc|cc}
\toprule
\multicolumn{1}{c|}{\multirow{2}{*}[-3.0em]{\textbf{Settings}}} &
\multicolumn{4}{c|}{\textbf{General}} &
\multicolumn{4}{c|}{\textbf{Text-rich}} &
\multicolumn{2}{c}{\textbf{Averages} $\uparrow$} \\
\cmidrule(lr){2-5}\cmidrule(lr){6-9}\cmidrule(lr){10-11}
&
\rot{MME} &
\rot{MMStar} &
\rot{RealWorldQA} &
\rot{SeedBench\textsubscript{image}} &
\rot{TextVQA} &
\rot{ChartQA\textsubscript{test}} &
\rot{OCRBench} &
\rot{DocVQA\textsubscript{test}} &
\rot{Gen-Avg} & \rot{Text-Avg} \\
\midrule

\multicolumn{11}{c}{\cellcolor{gray!20}\textbf{(a) Effect of Ratio}} \\
\midrule
Ratio = 0         & 1986 & 44.4 & \textbf{60.5} & 68.5 & 71.0 & 73.1 & 72.6 & 74.2 & 63.2 & 72.7 \\
Ratio = 1/8       & 2008 & 44.1 & 59.3 & 69.1 & 73.7 & 75.8 & 76.4 & 84.0 & 63.2 & 77.5 \\
\rowcolor{blue!10}
Ratio = 1/4       & \textbf{2010} & 45.0 & 60.0 & \textbf{70.6} & 76.0 & 77.5 & 78.6 & 87.8 & \textbf{64.2} & 80.0 \\
Ratio = full attn & 2008 & \textbf{45.1} & 59.8 & 70.5 & \textbf{76.2} & \textbf{77.6} & \textbf{78.8} & \textbf{88.1} & \textbf{63.9} & \textbf{80.2} \\
\midrule

\multicolumn{11}{c}{\cellcolor{gray!20}\textbf{(b) Effect of Architecture}} \\
\midrule
Linear attention & NAN & NAN & NAN & NAN & NAN & NAN & NAN & NAN & NAN & NAN \\
Mamba            & 1686 & 40.1 & 58.2 & 63.1 & 34.3 & 43.2 & 49.0 & 15.3 & 57.2 & 35.5 \\
GLA              & 1712 & 40.3 & 58.5 & 64.3 & 35.0 & 46.6 & 51.0 & 19.0 & 57.9 & 37.9 \\
\rowcolor{blue!10}
Gated DeltaNet   & \textbf{1986} & \textbf{44.4} & \textbf{60.5} & \textbf{68.5} & \textbf{71.0} & \textbf{73.1} & \textbf{72.6} & \textbf{74.2} & \textbf{63.2} & \textbf{72.7} \\
\midrule

\multicolumn{11}{c}{\cellcolor{gray!20}\textbf{(c) Effect of Training Stages}} \\
\midrule
None              & NAN & NAN & NAN & NAN & NAN & NAN & NAN & NAN & NAN & NAN \\
Stage 1           & 2010 & 45.0 & 60.0 & 70.6 & 76.0 & 77.5 & 78.6 & 87.8 & 64.0 & 80.0 \\
Stage 2           & 1713 & 39.6 & 54.1 & 65.7 & 73.4 & 73.0 & 74.6 & 84.1 & 57.0 & 76.3 \\
\rowcolor{blue!10}
Stage 1 + 2       & \textbf{2120} & \textbf{54.7} & \textbf{66.3} & \textbf{72.7} & \textbf{78.3} & \textbf{82.4} & \textbf{79.8} & \textbf{91.1} & \textbf{69.6} & \textbf{82.9} \\
\bottomrule
\end{tabular}}
\vspace{0.1cm}
\caption{\textbf{Ablation Studies.} Quantitative analysis validating the design choices of \texttt{InfiniteVL}, including (a) the optimal mixing ratio of Full Attention, (b) the selection of the core linear modeling architecture, and (c) the necessity of the multi-stage training strategy.}
\label{tab:hybrid_ablation_main}
\end{table*}

\subsection{Ablation Studies} 
\label{sec:ablation}  
We systematically analyze the design choices and training strategies of InfiniteVL-Base through ablation studies, focusing on the linear layer selection, mixing ratios, and the multi-stage training strategy (summarized in \cref{tab:hybrid_ablation_main}).
 
\subsubsection{Linear Layer Selection.} We evaluate the impact of different linear sequence modeling modules on VLM performance. Vanilla Linear Attention exhibits severe training instability~\cite{katharopoulos2020transformers} and fails to inherit the teacher model's pre-trained knowledge. While models incorporating scalar gating mechanisms (e.g., Mamba~\cite{gu2024mamba} and GLA~\cite{yanggla}) achieve convergence, they show limited performance on information-dense document and text understanding tasks. In contrast, Gated DeltaNet~\cite{yanggated}, with its highly efficient state compression mechanism, yields significant improvements in both training stability and downstream performance (particularly on DocVQA and OCRBench), proving that efficient state compression is crucial for precise visual tasks.       
   
\subsubsection{Mixing Strategy.} We investigate the impact of the Full Attention layer ratio within the architecture. Even a minimal proportion of Attention layers significantly boosts performance on visual intensive benchmarks. When the ratio reaches 1:3 (one Attention layer for every three linear layers), the hybrid architecture achieves performance parity with the Transformer baseline across comprehensive benchmarks.          

\subsubsection{Training Strategy.} Direct initialization of the linear layers fails to produce usable capabilities. Stage I (Distillation Pre-training) successfully imparts foundational conversational abilities. Following Stage II (Continuous SFT), the model surpasses the teacher model across multiple metrics, indicating superior generalization. Notably, skipping Stage I and training from scratch using only SFT yields extremely poor results, underscoring the absolute necessity of the distillation phase. Finally, as illustrated in \cref{tab:longcontext}, InfiniteVL-Base without long-context fine-tuning (InfiniteVL-Base (wo LSFT)) falls significantly behind its fine-tuned counterparts on multimodal long-context tasks, firmly validating the effectiveness of the Stage III fine-tuning for extended multimodal understanding.

\section{Conclusion}
\label{sec:conclusion}

In this paper, we introduced \textbf{InfiniteVL}, a novel hybrid vision-language architecture designed to reconcile the conflict between precise visual perception and long-context efficiency. By strategically interleaving constant-memory Gated DeltaNet layers with a small fraction of Full Attention layers, InfiniteVL-Base effectively overcomes the detail-degradation issues inherent in linear models. To further scale to extremely long or continuous multimodal contexts, we proposed a seamless long-sequence architectural fine-tuning strategy. This yields two specialized variants: InfiniteVL-Offline for precise offline long-video comprehension, and InfiniteVL-Online for constant-memory online streaming. Experiments demonstrate that InfiniteVL matches the capabilities of strictly controlled Transformer baselines, whilst delivering a 5$\times$ prefill speedup at 256K context lengths and enabling stable, real-time streaming at 24 FPS. Ultimately, InfiniteVL paves a highly efficient and scalable path for deploying robust, continuous multimodal agents in real-world environments.

{
    \small
    \bibliographystyle{ieeenat_fullname}
    \bibliography{main}

\begin{thebibliography}{83}
\providecommand{\natexlab}[1]{#1}
\providecommand{\url}[1]{\texttt{#1}}
\expandafter\ifx\csname urlstyle\endcsname\relax
  \providecommand{\doi}[1]{doi: #1}\else
  \providecommand{\doi}{doi: \begingroup \urlstyle{rm}\Url}\fi

\bibitem[Ainslie et~al.(2023)Ainslie, Lee-Thorp, de~Jong, Zemlyanskiy, Lebron, and Sanghai]{ainslie2023gqa}
Joshua Ainslie, James Lee-Thorp, Michiel de Jong, Yury Zemlyanskiy, Federico Lebron, and Sumit Sanghai.
\newblock Gqa: Training generalized multi-query transformer models from multi-head checkpoints.
\newblock In \emph{Proceedings of the 2023 Conference on Empirical Methods in Natural Language Processing}, pages 4895--4901, 2023.

\bibitem[An et~al.(2025)An, Xie, Yang, Zhang, Zhao, Cheng, Wang, Xu, Chen, Wu, et~al.]{an2025llava}
Xiang An, Yin Xie, Kaicheng Yang, Wenkang Zhang, Xiuwei Zhao, Zheng Cheng, Yirui Wang, Songcen Xu, Changrui Chen, Chunsheng Wu, et~al.
\newblock Llava-onevision-1.5: Fully open framework for democratized multimodal training.
\newblock \emph{arXiv preprint arXiv:2509.23661}, 2025.

\bibitem[Bai et~al.(2025)Bai, Chen, Liu, Wang, Ge, Song, Dang, Wang, Wang, Tang, et~al.]{bai2025qwen2}
Shuai Bai, Keqin Chen, Xuejing Liu, Jialin Wang, Wenbin Ge, Sibo Song, Kai Dang, Peng Wang, Shijie Wang, Jun Tang, et~al.
\newblock Qwen2. 5-vl technical report.
\newblock \emph{arXiv preprint arXiv:2502.13923}, 2025.

\bibitem[Black et~al.()Black, Brown, Driess, Esmail, Equi, Finn, Fusai, Groom, Hausman, Ichter, et~al.]{black2410pi0}
Kevin Black, Noah Brown, Danny Driess, Adnan Esmail, Michael Equi, Chelsea Finn, Niccolo Fusai, Lachy Groom, Karol Hausman, Brian Ichter, et~al.
\newblock $\pi$0: A vision-language-action flow model for general robot control. corr, abs/2410.24164, 2024. doi: 10.48550.
\newblock \emph{arXiv preprint ARXIV.2410.24164}.

\bibitem[Blakeman et~al.(2025)Blakeman, Basant, Khattar, Renduchintala, Bercovich, Ficek, Bjorlin, Taghibakhshi, Deshmukh, Mahabaleshwarkar, et~al.]{blakeman2025nemotron}
Aaron Blakeman, Aarti Basant, Abhinav Khattar, Adithya Renduchintala, Akhiad Bercovich, Aleksander Ficek, Alexis Bjorlin, Ali Taghibakhshi, Amala~Sanjay Deshmukh, Ameya~Sunil Mahabaleshwarkar, et~al.
\newblock Nemotron-h: A family of accurate and efficient hybrid mamba-transformer models.
\newblock \emph{arXiv preprint arXiv:2504.03624}, 2025.

\bibitem[Brown et~al.(2020)Brown, Mann, Ryder, Subbiah, Kaplan, Dhariwal, Neelakantan, Shyam, Sastry, Askell, et~al.]{brown2020language}
Tom Brown, Benjamin Mann, Nick Ryder, Melanie Subbiah, Jared~D Kaplan, Prafulla Dhariwal, Arvind Neelakantan, Pranav Shyam, Girish Sastry, Amanda Askell, et~al.
\newblock Language models are few-shot learners.
\newblock \emph{Advances in neural information processing systems}, 33:\penalty0 1877--1901, 2020.

\bibitem[Chen et~al.(2025)Chen, Li, Gong, Jiang, Fei, Yang, Shan, Yu, Wang, Zhu, et~al.]{chen2025minimax}
Aili Chen, Aonian Li, Bangwei Gong, Binyang Jiang, Bo Fei, Bo Yang, Boji Shan, Changqing Yu, Chao Wang, Cheng Zhu, et~al.
\newblock Minimax-m1: Scaling test-time compute efficiently with lightning attention.
\newblock \emph{arXiv preprint arXiv:2506.13585}, 2025.

\bibitem[Chen et~al.(2024{\natexlab{a}})Chen, Li, Dong, Zhang, Zang, Chen, Duan, Wang, Qiao, Lin, et~al.]{chen2024we}
Lin Chen, Jinsong Li, Xiaoyi Dong, Pan Zhang, Yuhang Zang, Zehui Chen, Haodong Duan, Jiaqi Wang, Yu Qiao, Dahua Lin, et~al.
\newblock Are we on the right way for evaluating large vision-language models?
\newblock \emph{Advances in Neural Information Processing Systems}, 37:\penalty0 27056--27087, 2024{\natexlab{a}}.

\bibitem[Chen et~al.(2024{\natexlab{b}})Chen, Zhao, Liu, Bai, Lin, Zhou, and Chang]{chen2024image}
Liang Chen, Haozhe Zhao, Tianyu Liu, Shuai Bai, Junyang Lin, Chang Zhou, and Baobao Chang.
\newblock An image is worth 1/2 tokens after layer 2: Plug-and-play inference acceleration for large vision-language models.
\newblock In \emph{European Conference on Computer Vision}, pages 19--35. Springer, 2024{\natexlab{b}}.

\bibitem[Chen et~al.()Chen, Xue, Li, Hu, Zhu, Li, Fang, Tang, Yang, Liu, et~al.]{chenlongvila}
Yukang Chen, Fuzhao Xue, Dacheng Li, Qinghao Hu, Ligeng Zhu, Xiuyu Li, Yunhao Fang, Haotian Tang, Shang Yang, Zhijian Liu, et~al.
\newblock Longvila: Scaling long-context visual language models for long videos.
\newblock In \emph{The Thirteenth International Conference on Learning Representations}.

\bibitem[Chen et~al.(2024{\natexlab{c}})Chen, Wang, Cao, Liu, Gao, Cui, Zhu, Ye, Tian, Liu, et~al.]{chen2024expanding}
Zhe Chen, Weiyun Wang, Yue Cao, Yangzhou Liu, Zhangwei Gao, Erfei Cui, Jinguo Zhu, Shenglong Ye, Hao Tian, Zhaoyang Liu, et~al.
\newblock Expanding performance boundaries of open-source multimodal models with model, data, and test-time scaling.
\newblock \emph{arXiv preprint arXiv:2412.05271}, 2024{\natexlab{c}}.

\bibitem[Dao(2023)]{dao2023flashattention}
Tri Dao.
\newblock Flashattention-2: Faster attention with better parallelism and work partitioning.
\newblock \emph{arXiv preprint arXiv:2307.08691}, 2023.

\bibitem[Deitke et~al.(2024)Deitke, Clark, Lee, Tripathi, Yang, Park, Salehi, Muennighoff, Lo, Soldaini, et~al.]{deitke2024molmo}
Matt Deitke, Christopher Clark, Sangho Lee, Rohun Tripathi, Yue Yang, Jae~Sung Park, Mohammadreza Salehi, Niklas Muennighoff, Kyle Lo, Luca Soldaini, et~al.
\newblock Molmo and pixmo: Open weights and open data for state-of-the-art multimodal models.
\newblock \emph{arXiv e-prints}, pages arXiv--2409, 2024.

\bibitem[Driess et~al.(2023)Driess, Xia, Sajjadi, Lynch, Chowdhery, Ichter, Wahid, Tompson, Vuong, Yu, et~al.]{driess2023palm}
Danny Driess, Fei Xia, Mehdi~SM Sajjadi, Corey Lynch, Aakanksha Chowdhery, Brian Ichter, Ayzaan Wahid, Jonathan Tompson, Quan Vuong, Tianhe Yu, et~al.
\newblock Palm-e: An embodied multimodal language model.
\newblock In \emph{International Conference on Machine Learning}, pages 8469--8488. PMLR, 2023.

\bibitem[Duan et~al.(2024)Duan, Yang, Qiao, Fang, Chen, Liu, Dong, Zang, Zhang, Wang, et~al.]{duan2024vlmevalkit}
Haodong Duan, Junming Yang, Yuxuan Qiao, Xinyu Fang, Lin Chen, Yuan Liu, Xiaoyi Dong, Yuhang Zang, Pan Zhang, Jiaqi Wang, et~al.
\newblock Vlmevalkit: An open-source toolkit for evaluating large multi-modality models.
\newblock In \emph{Proceedings of the 32nd ACM international conference on multimedia}, pages 11198--11201, 2024.

\bibitem[Fu et~al.(2025{\natexlab{a}})Fu, Chen, Shen, Qin, Zhang, Lin, Yang, Zheng, Li, Sun, et~al.]{fu2025mme}
Chaoyou Fu, Peixian Chen, Yunhang Shen, Yulei Qin, Mengdan Zhang, Xu Lin, Jinrui Yang, Xiawu Zheng, Ke Li, Xing Sun, et~al.
\newblock Mme: A comprehensive evaluation benchmark for multimodal large language models.
\newblock In \emph{The Thirty-ninth Annual Conference on Neural Information Processing Systems Datasets and Benchmarks Track}, 2025{\natexlab{a}}.

\bibitem[Fu et~al.(2025{\natexlab{b}})Fu, Dai, Luo, Li, Ren, Zhang, Wang, Zhou, Shen, Zhang, et~al.]{fu2025video}
Chaoyou Fu, Yuhan Dai, Yongdong Luo, Lei Li, Shuhuai Ren, Renrui Zhang, Zihan Wang, Chenyu Zhou, Yunhang Shen, Mengdan Zhang, et~al.
\newblock Video-mme: The first-ever comprehensive evaluation benchmark of multi-modal llms in video analysis.
\newblock In \emph{Proceedings of the Computer Vision and Pattern Recognition Conference}, pages 24108--24118, 2025{\natexlab{b}}.

\bibitem[Fu et~al.(2024)Fu, Hu, Li, Feng, Wang, Lin, Roth, Smith, Ma, and Krishna]{fu2024blink}
Xingyu Fu, Yushi Hu, Bangzheng Li, Yu Feng, Haoyu Wang, Xudong Lin, Dan Roth, Noah~A Smith, Wei-Chiu Ma, and Ranjay Krishna.
\newblock Blink: Multimodal large language models can see but not perceive.
\newblock In \emph{European Conference on Computer Vision}, pages 148--166. Springer, 2024.

\bibitem[Glorioso et~al.(2024)Glorioso, Anthony, Tokpanov, Whittington, Pilault, Ibrahim, and Millidge]{glorioso2024zamba}
Paolo Glorioso, Quentin Anthony, Yury Tokpanov, James Whittington, Jonathan Pilault, Adam Ibrahim, and Beren Millidge.
\newblock Zamba: A compact 7b ssm hybrid model.
\newblock \emph{arXiv preprint arXiv:2405.16712}, 2024.

\bibitem[{Google DeepMind and Google Research}(2025)]{google_pali_gemma2_modelcard_2025}
{Google DeepMind and Google Research}.
\newblock Paligemma 2 — google ai for developers (model page).
\newblock \url{https://ai.google.dev/gemma/docs/paligemma}, 2025.
\newblock Accessed 2025-11-12.

\bibitem[Gu and Dao()]{gu2024mamba}
Albert Gu and Tri Dao.
\newblock Mamba: Linear-time sequence modeling with selective state spaces.
\newblock In \emph{First conference on language modeling}.

\bibitem[Guo et~al.(2025)Guo, Wu, Zhu, Leng, Shi, Chen, Fan, Wang, Jiang, Wang, et~al.]{guo2025seed1}
Dong Guo, Faming Wu, Feida Zhu, Fuxing Leng, Guang Shi, Haobin Chen, Haoqi Fan, Jian Wang, Jianyu Jiang, Jiawei Wang, et~al.
\newblock Seed1. 5-vl technical report.
\newblock \emph{arXiv preprint arXiv:2505.07062}, 2025.

\bibitem[Hou et~al.(2025)Hou, Zeng, Ma, and Yu]{hou2025visualrwkv}
Haowen Hou, Peigen Zeng, Fei Ma, and Fei~Richard Yu.
\newblock Visualrwkv: Exploring recurrent neural networks for visual language models.
\newblock In \emph{Proceedings of the 31st International Conference on Computational Linguistics}, pages 10423--10434, 2025.

\bibitem[Jiang et~al.(2024)Jiang, Chen, Liao, Zhang, Yin, Zhang, Huang, Liu, and Wang]{jiang2024senna}
Bo Jiang, Shaoyu Chen, Bencheng Liao, Xingyu Zhang, Wei Yin, Qian Zhang, Chang Huang, Wenyu Liu, and Xinggang Wang.
\newblock Senna: Bridging large vision-language models and end-to-end autonomous driving.
\newblock \emph{arXiv preprint arXiv:2410.22313}, 2024.

\bibitem[Katharopoulos et~al.(2020)Katharopoulos, Vyas, Pappas, and Fleuret]{katharopoulos2020transformers}
Angelos Katharopoulos, Apoorv Vyas, Nikolaos Pappas, and Fran{\c{c}}ois Fleuret.
\newblock Transformers are rnns: Fast autoregressive transformers with linear attention.
\newblock In \emph{International conference on machine learning}, pages 5156--5165. PMLR, 2020.

\bibitem[Kembhavi et~al.(2016)Kembhavi, Salvato, Kolve, Seo, Hajishirzi, and Farhadi]{kembhavi2016diagram}
Aniruddha Kembhavi, Mike Salvato, Eric Kolve, Minjoon Seo, Hannaneh Hajishirzi, and Ali Farhadi.
\newblock A diagram is worth a dozen images.
\newblock In \emph{European conference on computer vision}, pages 235--251. Springer, 2016.

\bibitem[Laurençon et~al.(2024{\natexlab{a}})Laurençon, Marafioti, Sanh, and Tronchon]{laurençon2024building}
Hugo Laurençon, Andrés Marafioti, Victor Sanh, and Léo Tronchon.
\newblock Building and better understanding vision-language models: insights and future directions., 2024{\natexlab{a}}.

\bibitem[Laurençon et~al.(2024{\natexlab{b}})Laurençon, Tronchon, Cord, and Sanh]{laurençon2024matters}
Hugo Laurençon, Léo Tronchon, Matthieu Cord, and Victor Sanh.
\newblock What matters when building vision-language models?, 2024{\natexlab{b}}.

\bibitem[Li et~al.(2024{\natexlab{a}})Li, Ge, Ge, Wang, Wang, Zhang, and Shan]{li2024seed}
Bohao Li, Yuying Ge, Yixiao Ge, Guangzhi Wang, Rui Wang, Ruimao Zhang, and Ying Shan.
\newblock Seed-bench: Benchmarking multimodal large language models.
\newblock In \emph{Proceedings of the IEEE/CVF Conference on Computer Vision and Pattern Recognition}, pages 13299--13308, 2024{\natexlab{a}}.

\bibitem[Li et~al.(2023)Li, Li, Savarese, and Hoi]{li2023blip}
Junnan Li, Dongxu Li, Silvio Savarese, and Steven Hoi.
\newblock Blip-2: Bootstrapping language-image pre-training with frozen image encoders and large language models.
\newblock In \emph{International conference on machine learning}, pages 19730--19742. PMLR, 2023.

\bibitem[Li et~al.(2024{\natexlab{b}})Li, Li, Wang, He, Wang, Wang, and Qiao]{li2024videomamba}
Kunchang Li, Xinhao Li, Yi Wang, Yinan He, Yali Wang, Limin Wang, and Yu Qiao.
\newblock Videomamba: State space model for efficient video understanding.
\newblock In \emph{European conference on computer vision}, pages 237--255. Springer, 2024{\natexlab{b}}.

\bibitem[Li et~al.(2025{\natexlab{a}})Li, Liao, Liu, and Wang]{li2025matvlm}
Yingyue Li, Bencheng Liao, Wenyu Liu, and Xinggang Wang.
\newblock Matvlm: Hybrid mamba-transformer for efficient vision-language modeling.
\newblock \emph{arXiv preprint arXiv:2503.13440}, 2025{\natexlab{a}}.

\bibitem[Li et~al.(2025{\natexlab{b}})Li, Xiong, Guo, Li, Yan, Xu, Zhou, Chen, Sun, Wang, et~al.]{li2025recogdrive}
Yongkang Li, Kaixin Xiong, Xiangyu Guo, Fang Li, Sixu Yan, Gangwei Xu, Lijun Zhou, Long Chen, Haiyang Sun, Bing Wang, et~al.
\newblock Recogdrive: A reinforced cognitive framework for end-to-end autonomous driving.
\newblock \emph{arXiv preprint arXiv:2506.08052}, 2025{\natexlab{b}}.

\bibitem[Liao et~al.(2025{\natexlab{a}})Liao, Tao, Zhang, Cheng, Li, Yin, Liu, and Wang]{liao2025multimodal}
Bencheng Liao, Hongyuan Tao, Qian Zhang, Tianheng Cheng, Yingyue Li, Haoran Yin, Wenyu Liu, and Xinggang Wang.
\newblock Multimodal mamba: Decoder-only multimodal state space model via quadratic to linear distillation.
\newblock \emph{arXiv preprint arXiv:2502.13145}, 2025{\natexlab{a}}.

\bibitem[Liao et~al.(2025{\natexlab{b}})Liao, Wang, Zhu, Zhang, and Huang]{vig}
Bencheng Liao, Xinggang Wang, Lianghui Zhu, Qian Zhang, and Chang Huang.
\newblock Vig: Linear-complexity visual sequence learning with gated linear attention.
\newblock In \emph{Proceedings of the AAAI Conference on Artificial Intelligence}, pages 5182--5190, 2025{\natexlab{b}}.

\bibitem[Lieber et~al.(2024)Lieber, Lenz, Bata, Cohen, Osin, Dalmedigos, Safahi, Meirom, Belinkov, Shalev-Shwartz, et~al.]{lieber2024jamba}
Opher Lieber, Barak Lenz, Hofit Bata, Gal Cohen, Jhonathan Osin, Itay Dalmedigos, Erez Safahi, Shaked Meirom, Yonatan Belinkov, Shai Shalev-Shwartz, et~al.
\newblock Jamba: A hybrid transformer-mamba language model.
\newblock \emph{arXiv preprint arXiv:2403.19887}, 2024.

\bibitem[Lin et~al.(2024{\natexlab{a}})Lin, Fang, Chen, Wan, Luo, Li, Liu, and Sun]{lin2024streamingbench}
Junming Lin, Zheng Fang, Chi Chen, Zihao Wan, Fuwen Luo, Peng Li, Yang Liu, and Maosong Sun.
\newblock Streamingbench: Assessing the gap for mllms to achieve streaming video understanding.
\newblock \emph{arXiv preprint arXiv:2411.03628}, 2024{\natexlab{a}}.

\bibitem[Lin et~al.(2024{\natexlab{b}})Lin, Yin, Ping, Molchanov, Shoeybi, and Han]{lin2024vila}
Ji Lin, Hongxu Yin, Wei Ping, Pavlo Molchanov, Mohammad Shoeybi, and Song Han.
\newblock Vila: On pre-training for visual language models.
\newblock In \emph{2024 IEEE/CVF Conference on Computer Vision and Pattern Recognition (CVPR)}, pages 26679--26689. IEEE, 2024{\natexlab{b}}.

\bibitem[Liu et~al.(2023)Liu, Li, Wu, and Lee]{liu2023visual}
Haotian Liu, Chunyuan Li, Qingyang Wu, and Yong~Jae Lee.
\newblock Visual instruction tuning.
\newblock \emph{Advances in neural information processing systems}, 36:\penalty0 34892--34916, 2023.

\bibitem[Liu et~al.(2024{\natexlab{a}})Liu, Duan, Zhang, Li, Zhang, Zhao, Yuan, Wang, He, Liu, et~al.]{liu2024mmbench}
Yuan Liu, Haodong Duan, Yuanhan Zhang, Bo Li, Songyang Zhang, Wangbo Zhao, Yike Yuan, Jiaqi Wang, Conghui He, Ziwei Liu, et~al.
\newblock Mmbench: Is your multi-modal model an all-around player?
\newblock In \emph{European conference on computer vision}, pages 216--233. Springer, 2024{\natexlab{a}}.

\bibitem[Liu et~al.(2024{\natexlab{b}})Liu, Li, Huang, Yang, Yu, Li, Yin, Liu, Jin, and Bai]{liu2024ocrbench}
Yuliang Liu, Zhang Li, Mingxin Huang, Biao Yang, Wenwen Yu, Chunyuan Li, Xu-Cheng Yin, Cheng-Lin Liu, Lianwen Jin, and Xiang Bai.
\newblock Ocrbench: on the hidden mystery of ocr in large multimodal models.
\newblock \emph{Science China Information Sciences}, 67\penalty0 (12):\penalty0 220102, 2024{\natexlab{b}}.

\bibitem[Lu et~al.(2023)Lu, Bansal, Xia, Liu, Li, Hajishirzi, Cheng, Chang, Galley, and Gao]{lu2023mathvista}
Pan Lu, Hritik Bansal, Tony Xia, Jiacheng Liu, Chunyuan Li, Hannaneh Hajishirzi, Hao Cheng, Kai-Wei Chang, Michel Galley, and Jianfeng Gao.
\newblock Mathvista: Evaluating math reasoning in visual contexts with gpt-4v, bard, and other large multimodal models.
\newblock \emph{CoRR}, 2023.

\bibitem[Maaz et~al.(2024)Maaz, Rasheed, Khan, and Khan]{maaz2024video}
Muhammad Maaz, Hanoona Rasheed, Salman Khan, and Fahad Khan.
\newblock Video-chatgpt: Towards detailed video understanding via large vision and language models.
\newblock In \emph{Proceedings of the 62nd Annual Meeting of the Association for Computational Linguistics (Volume 1: Long Papers)}, pages 12585--12602, 2024.

\bibitem[Marafioti et~al.(2025)Marafioti, Zohar, Farré, Noyan, Bakouch, Cuenca, Zakka, Allal, Lozhkov, Tazi, Srivastav, Lochner, Larcher, Morlon, Tunstall, von Werra, and Wolf]{marafioti2025smolvlm}
Andrés Marafioti, Orr Zohar, Miquel Farré, Merve Noyan, Elie Bakouch, Pedro Cuenca, Cyril Zakka, Loubna~Ben Allal, Anton Lozhkov, Nouamane Tazi, Vaibhav Srivastav, Joshua Lochner, Hugo Larcher, Mathieu Morlon, Lewis Tunstall, Leandro von Werra, and Thomas Wolf.
\newblock Smolvlm: Redefining small and efficient multimodal models.
\newblock \emph{arXiv preprint arXiv:2504.05299}, 2025.

\bibitem[Masry et~al.(2022)Masry, Do, Tan, Joty, and Hoque]{masry2022chartqa}
Ahmed Masry, Xuan~Long Do, Jia~Qing Tan, Shafiq Joty, and Enamul Hoque.
\newblock Chartqa: A benchmark for question answering about charts with visual and logical reasoning.
\newblock In \emph{Findings of the association for computational linguistics: ACL 2022}, pages 2263--2279, 2022.

\bibitem[Mathew et~al.(2021)Mathew, Karatzas, and Jawahar]{mathew2021docvqa}
Minesh Mathew, Dimosthenis Karatzas, and CV Jawahar.
\newblock Docvqa: A dataset for vqa on document images.
\newblock In \emph{Proceedings of the IEEE/CVF winter conference on applications of computer vision}, pages 2200--2209, 2021.

\bibitem[{Microsoft}(2025)]{microsoft_phi35vision_hf_2025}
{Microsoft}.
\newblock Phi-3.5-vision-instruct (model card).
\newblock \url{https://huggingface.co/microsoft/Phi-3.5-vision-instruct}, 2025.
\newblock Lightweight multimodal model with 128K context; official model card. Accessed 2025-11-12.

\bibitem[Pozzi et~al.(2025)Pozzi, Incremona, Tessera, and Toti]{pozzi2025mitigating}
Andrea Pozzi, Alessandro Incremona, Daniele Tessera, and Daniele Toti.
\newblock Mitigating exposure bias in large language model distillation: an imitation learning approach.
\newblock \emph{Neural Computing and Applications}, pages 1--17, 2025.

\bibitem[Qiao et~al.(2024)Qiao, Yu, Zhao, Chen, Sun, Guo, Wu, and Liu]{qiao2024vl}
Yanyuan Qiao, Zheng Yu, Zijia Zhao, Sihan Chen, Mingzhen Sun, Longteng Guo, Qi Wu, and Jing Liu.
\newblock Vl-mamba: Exploring state space models for multimodal learning.
\newblock In \emph{NeurIPS Efficient Natural Language and Speech Processing Workshop}, pages 102--113. PMLR, 2024.

\bibitem[Ren et~al.(2024)Ren, Liu, Lu, Shen, Liang, and Chen]{ren2024samba}
Liliang Ren, Yang Liu, Yadong Lu, Yelong Shen, Chen Liang, and Weizhu Chen.
\newblock Samba: Simple hybrid state space models for efficient unlimited context language modeling.
\newblock \emph{arXiv preprint arXiv:2406.07522}, 2024.

\bibitem[Saikh et~al.(2022)Saikh, Ghosal, Mittal, Ekbal, and Bhattacharyya]{saikh2022scienceqa}
Tanik Saikh, Tirthankar Ghosal, Amish Mittal, Asif Ekbal, and Pushpak Bhattacharyya.
\newblock Scienceqa: A novel resource for question answering on scholarly articles.
\newblock \emph{International Journal on Digital Libraries}, 23\penalty0 (3):\penalty0 289--301, 2022.

\bibitem[Schlag et~al.(2021)Schlag, Irie, and Schmidhuber]{schlag2021linear}
Imanol Schlag, Kazuki Irie, and J{\"u}rgen Schmidhuber.
\newblock Linear transformers are secretly fast weight programmers.
\newblock In \emph{International conference on machine learning}, pages 9355--9366. PMLR, 2021.

\bibitem[Singh et~al.(2019)Singh, Natarajan, Shah, Jiang, Chen, Batra, Parikh, and Rohrbach]{singh2019towards}
Amanpreet Singh, Vivek Natarajan, Meet Shah, Yu Jiang, Xinlei Chen, Dhruv Batra, Devi Parikh, and Marcus Rohrbach.
\newblock Towards vqa models that can read.
\newblock In \emph{Proceedings of the IEEE/CVF conference on computer vision and pattern recognition}, pages 8317--8326, 2019.

\bibitem[Song et~al.(2024)Song, Chai, Wang, Zhang, Zhou, Wu, Chi, Guo, Ye, Zhang, et~al.]{song2024moviechat}
Enxin Song, Wenhao Chai, Guanhong Wang, Yucheng Zhang, Haoyang Zhou, Feiyang Wu, Haozhe Chi, Xun Guo, Tian Ye, Yanting Zhang, et~al.
\newblock Moviechat: From dense token to sparse memory for long video understanding.
\newblock In \emph{Proceedings of the IEEE/CVF Conference on Computer Vision and Pattern Recognition}, pages 18221--18232, 2024.

\bibitem[Song et~al.(2025)Song, Chai, Yang, Armand, Shan, Xu, Xie, and Tu]{song2025videonsa}
Enxin Song, Wenhao Chai, Shusheng Yang, Ethan Armand, Xiaojun Shan, Haiyang Xu, Jianwen Xie, and Zhuowen Tu.
\newblock Videonsa: Native sparse attention scales video understanding.
\newblock \emph{arXiv preprint arXiv:2510.02295}, 2025.

\bibitem[Su et~al.(2024)Su, Ahmed, Lu, Pan, Bo, and Liu]{su2024roformer}
Jianlin Su, Murtadha Ahmed, Yu Lu, Shengfeng Pan, Wen Bo, and Yunfeng Liu.
\newblock Roformer: Enhanced transformer with rotary position embedding.
\newblock \emph{Neurocomputing}, 568:\penalty0 127063, 2024.

\bibitem[Team et~al.(2025)Team, Zhang, Lin, Yao, Hu, Meng, Liu, Men, Yang, Li, et~al.]{team2025kimi}
Kimi Team, Yu Zhang, Zongyu Lin, Xingcheng Yao, Jiaxi Hu, Fanqing Meng, Chengyin Liu, Xin Men, Songlin Yang, Zhiyuan Li, et~al.
\newblock Kimi linear: An expressive, efficient attention architecture.
\newblock \emph{arXiv preprint arXiv:2510.26692}, 2025.

\bibitem[Team et~al.(2026)Team, An, Chen, Fang, Li, Li, Li, Li, Li, Lin, et~al.]{team2026minicpm}
MiniCPM Team, Wenhao An, Yingfa Chen, Yewei Fang, Jiayi Li, Xin Li, Yaohui Li, Yishan Li, Yuxuan Li, Biyuan Lin, et~al.
\newblock Minicpm-sala: Hybridizing sparse and linear attention for efficient long-context modeling.
\newblock \emph{arXiv preprint arXiv:2602.11761}, 2026.

\bibitem[Touvron et~al.(2023)Touvron, Lavril, Izacard, Martinet, Lachaux, Lacroix, Rozi{\`e}re, Goyal, Hambro, Azhar, et~al.]{touvron2023llama}
Hugo Touvron, Thibaut Lavril, Gautier Izacard, Xavier Martinet, Marie-Anne Lachaux, Timoth{\'e}e Lacroix, Baptiste Rozi{\`e}re, Naman Goyal, Eric Hambro, Faisal Azhar, et~al.
\newblock Llama: Open and efficient foundation language models.
\newblock \emph{arXiv preprint arXiv:2302.13971}, 2023.

\bibitem[Vaswani et~al.(2017)Vaswani, Shazeer, Parmar, Uszkoreit, Jones, Gomez, Kaiser, and Polosukhin]{vaswani2017attention}
Ashish Vaswani, Noam Shazeer, Niki Parmar, Jakob Uszkoreit, Llion Jones, Aidan~N Gomez, {\L}ukasz Kaiser, and Illia Polosukhin.
\newblock Attention is all you need.
\newblock \emph{Advances in neural information processing systems}, 30, 2017.

\bibitem[Wang et~al.(2025)Wang, Ding, Zeng, Zhou, Shen, Luo, Yu, and Tao]{wang2025divide}
Wenbin Wang, Liang Ding, Minyan Zeng, Xiabin Zhou, Li Shen, Yong Luo, Wei Yu, and Dacheng Tao.
\newblock Divide, conquer and combine: A training-free framework for high-resolution image perception in multimodal large language models.
\newblock In \emph{Proceedings of the AAAI Conference on Artificial Intelligence}, pages 7907--7915, 2025.

\bibitem[Wiedmann et~al.(2025)Wiedmann, Zohar, Mahla, Wang, Li, Frere, von Werra, Gosthipaty, and Marafioti]{wiedmann2025finevision}
Luis Wiedmann, Orr Zohar, Amir Mahla, Xiaohan Wang, Rui Li, Thibaud Frere, Leandro von Werra, Aritra~Roy Gosthipaty, and Andr{\'e}s Marafioti.
\newblock Finevision: Open data is all you need.
\newblock \emph{arXiv preprint arXiv:2510.17269}, 2025.

\bibitem[Wu et~al.(2024)Wu, Li, Chen, and Li]{wu2024longvideobench}
Haoning Wu, Dongxu Li, Bei Chen, and Junnan Li.
\newblock Longvideobench: A benchmark for long-context interleaved video-language understanding.
\newblock \emph{Advances in Neural Information Processing Systems}, 37:\penalty0 28828--28857, 2024.

\bibitem[Wu and Xie(2024)]{wu2024v}
Penghao Wu and Saining Xie.
\newblock V?: Guided visual search as a core mechanism in multimodal llms.
\newblock In \emph{Proceedings of the IEEE/CVF Conference on Computer Vision and Pattern Recognition}, pages 13084--13094, 2024.

\bibitem[xAI(2024)]{xai_realworldqa_2024}
xAI.
\newblock Realworldqa, 2024.

\bibitem[Xiao et~al.()Xiao, Tian, Chen, Han, and Lewis]{xiaoefficient}
Guangxuan Xiao, Yuandong Tian, Beidi Chen, Song Han, and Mike Lewis.
\newblock Efficient streaming language models with attention sinks.
\newblock In \emph{The Twelfth International Conference on Learning Representations}.

\bibitem[Xiao et~al.(2023)Xiao, Tian, Chen, Han, and Lewis]{xiao2023efficient}
Guangxuan Xiao, Yuandong Tian, Beidi Chen, Song Han, and Mike Lewis.
\newblock Efficient streaming language models with attention sinks.
\newblock \emph{arXiv preprint arXiv:2309.17453}, 2023.

\bibitem[Xu et~al.(2025)Xu, Xiao, Chen, He, Peng, Lu, and Han]{xu2025streamingvlm}
Ruyi Xu, Guangxuan Xiao, Yukang Chen, Liuning He, Kelly Peng, Yao Lu, and Song Han.
\newblock Streamingvlm: Real-time understanding for infinite video streams.
\newblock \emph{arXiv preprint arXiv:2510.09608}, 2025.

\bibitem[Yang et~al.({\natexlab{a}})Yang, Kautz, and Hatamizadeh]{yanggated}
Songlin Yang, Jan Kautz, and Ali Hatamizadeh.
\newblock Gated delta networks: Improving mamba2 with delta rule.
\newblock In \emph{The Thirteenth International Conference on Learning Representations}, {\natexlab{a}}.

\bibitem[Yang et~al.({\natexlab{b}})Yang, Wang, Shen, Panda, and Kim]{yanggla}
Songlin Yang, Bailin Wang, Yikang Shen, Rameswar Panda, and Yoon Kim.
\newblock Gated linear attention transformers with hardware-efficient training.
\newblock In \emph{Forty-first International Conference on Machine Learning}, {\natexlab{b}}.

\bibitem[Yang et~al.(2024)Yang, Wang, Zhang, Shen, and Kim]{yang2024parallelizing}
Songlin Yang, Bailin Wang, Yu Zhang, Yikang Shen, and Yoon Kim.
\newblock Parallelizing linear transformers with the delta rule over sequence length.
\newblock \emph{Advances in neural information processing systems}, 37:\penalty0 115491--115522, 2024.

\bibitem[Yang et~al.(2025)Yang, Yang, Huang, Brown, Yang, Yu, Tong, Zheng, Xu, Wang, et~al.]{yang2025cambrian}
Shusheng Yang, Jihan Yang, Pinzhi Huang, Ellis Brown, Zihao Yang, Yue Yu, Shengbang Tong, Zihan Zheng, Yifan Xu, Muhan Wang, et~al.
\newblock Cambrian-s: Towards spatial supersensing in video.
\newblock \emph{arXiv preprint arXiv:2511.04670}, 2025.

\bibitem[Yue et~al.(2024)Yue, Ni, Zhang, Zheng, Liu, Zhang, Stevens, Jiang, Ren, Sun, et~al.]{yue2024mmmu}
Xiang Yue, Yuansheng Ni, Kai Zhang, Tianyu Zheng, Ruoqi Liu, Ge Zhang, Samuel Stevens, Dongfu Jiang, Weiming Ren, Yuxuan Sun, et~al.
\newblock Mmmu: A massive multi-discipline multimodal understanding and reasoning benchmark for expert agi.
\newblock In \emph{Proceedings of the IEEE/CVF Conference on Computer Vision and Pattern Recognition}, pages 9556--9567, 2024.

\bibitem[Zeng et~al.(2025)Zeng, Yao, Liao, Tao, Liu, and Wang]{zeng2025diffusionvl}
Lunbin Zeng, Jingfeng Yao, Bencheng Liao, Hongyuan Tao, Wenyu Liu, and Xinggang Wang.
\newblock Diffusionvl: Translating any autoregressive models into diffusion vision language models.
\newblock \emph{arXiv preprint arXiv:2512.15713}, 2025.

\bibitem[Zhang et~al.(2024)Zhang, Wu, Li, Li, Ma, Liu, and Li]{zhang2024llava}
Yuanhan Zhang, Jinming Wu, Wei Li, Bo Li, Zejun Ma, Ziwei Liu, and Chunyuan Li.
\newblock Llava-video: Video instruction tuning with synthetic data.
\newblock \emph{arXiv preprint arXiv:2410.02713}, 2024.

\bibitem[Zhang et~al.(2023)Zhang, Sheng, Zhou, Chen, Zheng, Cai, Song, Tian, R{\'e}, Barrett, et~al.]{zhang2023h2o}
Zhenyu Zhang, Ying Sheng, Tianyi Zhou, Tianlong Chen, Lianmin Zheng, Ruisi Cai, Zhao Song, Yuandong Tian, Christopher R{\'e}, Clark Barrett, et~al.
\newblock H2o: Heavy-hitter oracle for efficient generative inference of large language models.
\newblock \emph{Advances in Neural Information Processing Systems}, 36:\penalty0 34661--34710, 2023.

\bibitem[Zhao et~al.(2025{\natexlab{a}})Zhao, Zhang, Zhao, Ding, Huang, and Wang]{zhao2025cobra}
Han Zhao, Min Zhang, Wei Zhao, Pengxiang Ding, Siteng Huang, and Donglin Wang.
\newblock Cobra: Extending mamba to multi-modal large language model for efficient inference.
\newblock In \emph{Proceedings of the AAAI Conference on Artificial Intelligence}, pages 10421--10429, 2025{\natexlab{a}}.

\bibitem[Zhao et~al.(2025{\natexlab{b}})Zhao, Zhou, Su, Xiao, Li, Li, Zhang, Zhao, Li, Huang, et~al.]{zhao2025infllm}
Weilin Zhao, Zihan Zhou, Zhou Su, Chaojun Xiao, Yuxuan Li, Yanghao Li, Yudi Zhang, Weilun Zhao, Zhen Li, Yuxiang Huang, et~al.
\newblock Infllm-v2: Dense-sparse switchable attention for seamless short-to-long adaptation.
\newblock \emph{arXiv preprint arXiv:2509.24663}, 2025{\natexlab{b}}.

\bibitem[Zhou et~al.(2024)Zhou, Hu, Weng, Jia, Luo, Liu, Wu, and Huang]{zhou2024tinyllava}
Baichuan Zhou, Ying Hu, Xi Weng, Junlong Jia, Jie Luo, Xien Liu, Ji Wu, and Lei Huang.
\newblock Tinyllava: A framework of small-scale large multimodal models.
\newblock \emph{arXiv preprint arXiv:2402.14289}, 2024.

\bibitem[Zhu et~al.()Zhu, Chen, Shen, Li, and Elhoseiny]{zhuminigpt}
Deyao Zhu, Jun Chen, Xiaoqian Shen, Xiang Li, and Mohamed Elhoseiny.
\newblock Minigpt-4: Enhancing vision-language understanding with advanced large language models.
\newblock In \emph{The Twelfth International Conference on Learning Representations}.

\bibitem[Zhu et~al.(2024)Zhu, Huang, Liao, Liew, Yan, Feng, and Wang]{dig}
Lianghui Zhu, Zilong Huang, Bencheng Liao, Jun~Hao Liew, Hanshu Yan, Jiashi Feng, and Xinggang Wang.
\newblock Dig: Scalable and efficient diffusion models with gated linear attention.
\newblock \emph{arXiv}, 2024.

\bibitem[Zitkovich et~al.(2023)Zitkovich, Yu, Xu, Xu, Xiao, Xia, Wu, Wohlhart, Welker, Wahid, et~al.]{zitkovich2023rt}
Brianna Zitkovich, Tianhe Yu, Sichun Xu, Peng Xu, Ted Xiao, Fei Xia, Jialin Wu, Paul Wohlhart, Stefan Welker, Ayzaan Wahid, et~al.
\newblock Rt-2: Vision-language-action models transfer web knowledge to robotic control.
\newblock In \emph{Conference on Robot Learning}, pages 2165--2183. PMLR, 2023.

\bibitem[Zou et~al.(2025)Zou, Liao, Zhang, Liu, and Wang]{zou2025omnimamba}
Jialv Zou, Bencheng Liao, Qian Zhang, Wenyu Liu, and Xinggang Wang.
\newblock Omnimamba: Efficient and unified multimodal understanding and generation via state space models.
\newblock \emph{arXiv preprint arXiv:2503.08686}, 2025.

\end{thebibliography}
}

\clearpage
\setcounter{page}{1}

\vspace*{1em} 
\noindent{\huge \bfseries \color{horizonblue} Appendix}
\par\vspace{1em} 

In this part, we provide additional details about InfiniteVL, which are omitted due to the 14 page limit of the main paper. Specifically, Section~\ref{sec:A} analyzes the cache evolution of linear layers in InfiniteVL under long-term streaming inputs, illustrating its length generalization capability. Section~\ref{sec:B} supplements comprehensive training and evaluation setup details to facilitate the reproduction of our experimental results. Section~\ref{sec:C} presents a series of  visualization  cases demonstrating InfiniteVL's performance in general VQA, text-rich understanding, and ultra-long streaming understanding scenarios.

\section{Analysis of Cache Evolution under Streaming Inputs}
\label{sec:A}
In the linear layers, each output is computed by multiplying the query with a memory cache formed through the cumulative outer product of keys and values. This memory cache, a matrix of size $16 \times 128 \times 256$, serves as one of the core components influencing the output, and its stability directly affects output stability. To long-term behavior in streaming scenarios, we track the evolution of the L2 norm of this memory cache as the number of input frames increases, using it as a metric for long-term stability. 

\begin{figure}[ht!]
    \centering
    \includegraphics[width=0.8\linewidth]{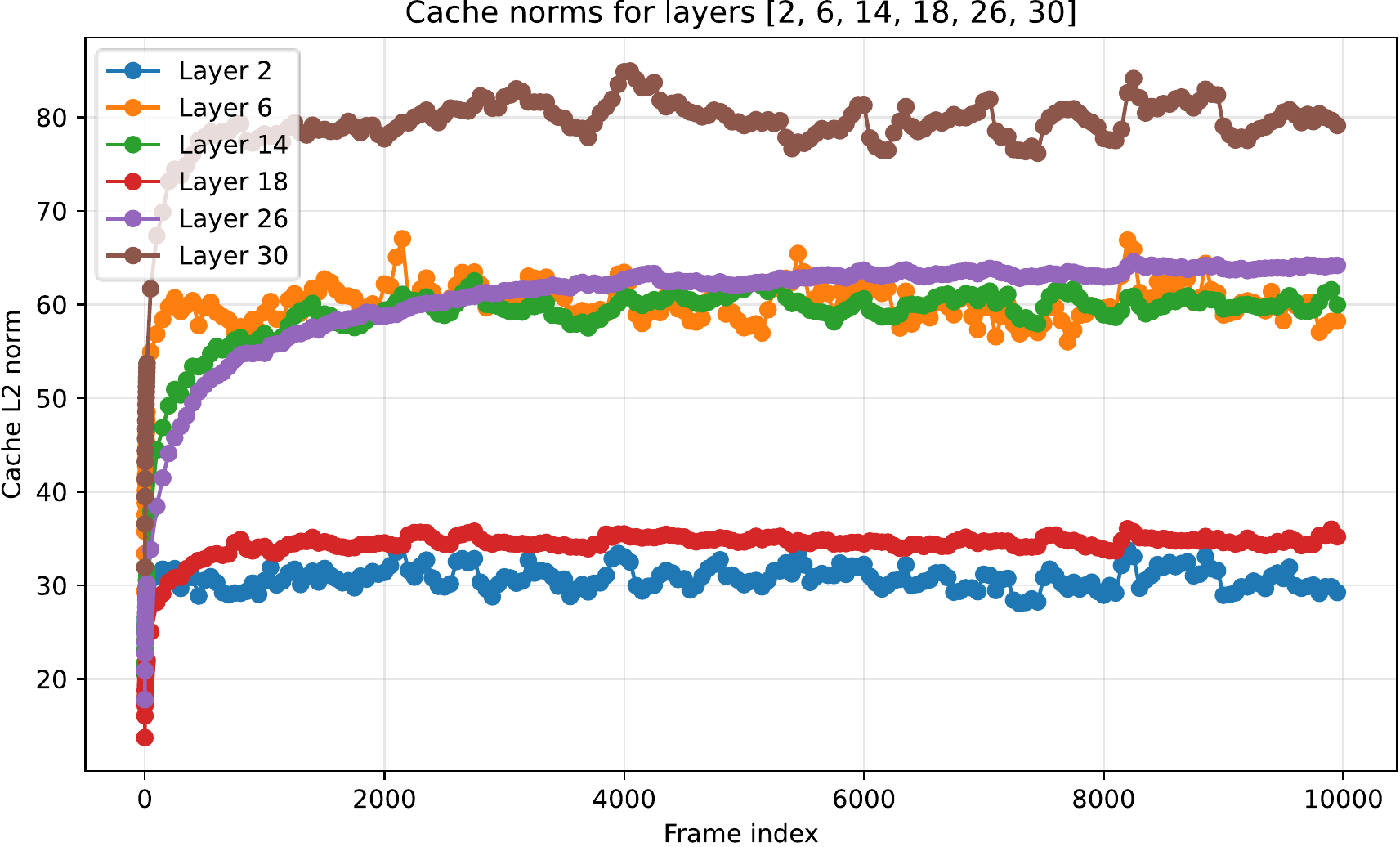}
    \caption{\textbf{L2 norm of the Linear-layer memory cache versus input frame index:} the norm increases rapidly at the beginning and then stabilizes.}
    \label{fig: cache_norm}
    \vspace{-0.3cm}
\end{figure}

As shown in Fig.~\ref{fig: cache_norm}, the norm of the memory cache rises rapidly during the initial frames and then produce convergence, demonstrating stable behavior without exhibiting unbounded growth. This indicates that the cache management mechanism remains effective and that the linear layers maintain stable comprehension capability when processing unlimited streaming input sequences.

\section{Comprehensive Training and Evaluation Setup Details}
\label{sec:B}
All trainings were conducted on NVIDIA H20 GPUs using bfloat16 (BF16) precision. The training corpus comprises exclusively publicly available, open-source datasets totaling $\sim$10M samples. For each stage, we uniformly subsample the required portion of data. We use AdamW with $\beta=(0.9,0.999)$ and weight decay $0.01$, along with a cosine-annealing learning-rate schedule and a 5\% linear warm-up.

In Stage~I, we use a learning rate of $2\times10^{-4}$, batch size of 64, with both layer-wise and end-to-end distillation trained on 1M multimodal Caption \& QA pairs each. Stage~II employs a learning rate of $5\times10^{-5}$, batch size of 256, and trains on 8M multimodal QA pairs. For Stage~III, the configuration includes a learning rate of $2\times10^{-5}$, batch size of 64, and training on 0.25M multimodal long-sequence Caption \& QA pairs together with 0.75M multimodal SFT data from Stage~II's training corpora to prevent excessive distribution shift.

We adopt \textbf{VLMEvalKit} as our multimodal evaluation framework and utilize its default prompt configurations for all benchmarks. During evaluation, the maximum generation length for InfiniteVL is set to 256 tokens, with image resolutions ranging from $128\times128$ to $1344\times1344$. By prepending all visual tokens to the prompt, we force the model to compress the full image without any textual priors, strictly validating InfiniteVL's highly efficient and lossless-like compression. For specialized tasks such as MMMU and MathVista-MINI, we employ Qwen3-4B-Instruct as the evaluation model to ensure consistent and reliable assessment of reasoning capabilities.

\section{Case Study across multiple scenarios}
\label{sec:C}
\begin{figure*}[ht!]
    \centering
    \includegraphics[width=1.0\linewidth]{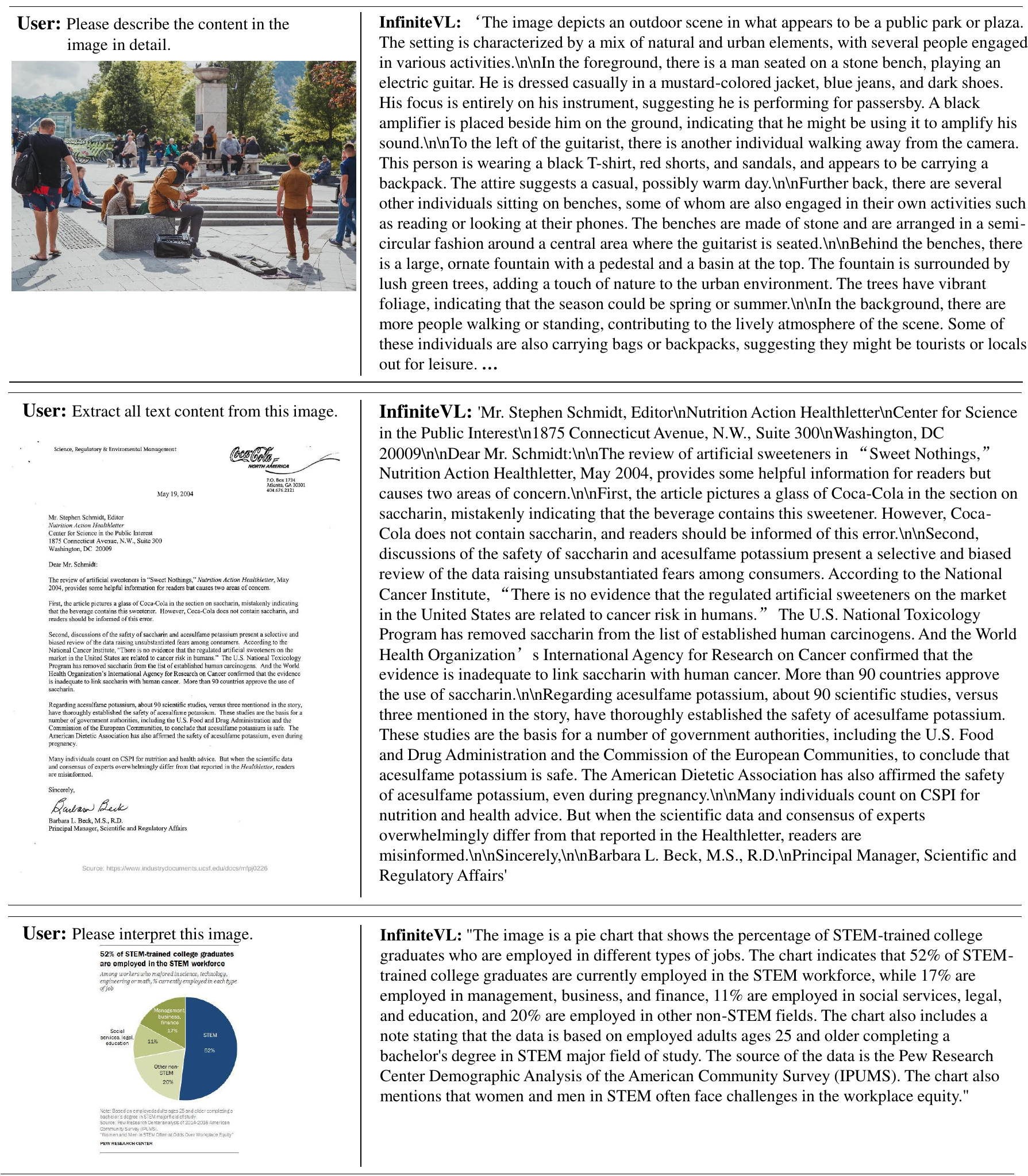}
    \vspace{-0.7cm}
    \caption{Examples of fundamental Visual-Language Understanding}
    \label{fig: image_case1}
    \vspace{-0.7cm}
\end{figure*}

\begin{figure*}[ht!]
    \centering
    \includegraphics[width=1.0\linewidth]{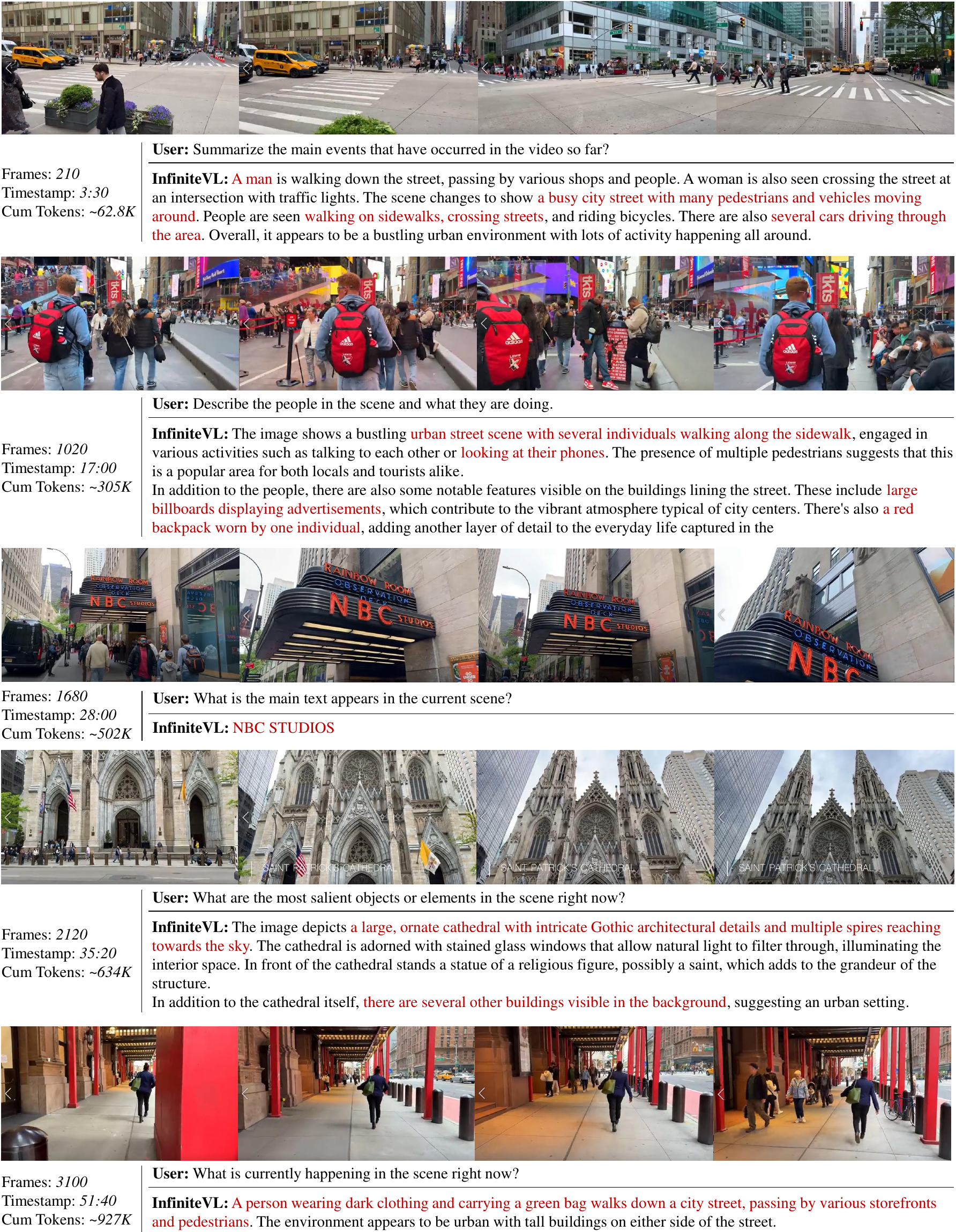}
    \vspace{-0.7cm}
    \caption{Examples of Long-Term Streaming Understanding Capability}
    \label{fig: streaming_case1}
    \vspace{-0.7cm}
\end{figure*}

\begin{figure*}[ht!]
    \centering
    \includegraphics[width=1.0\linewidth]{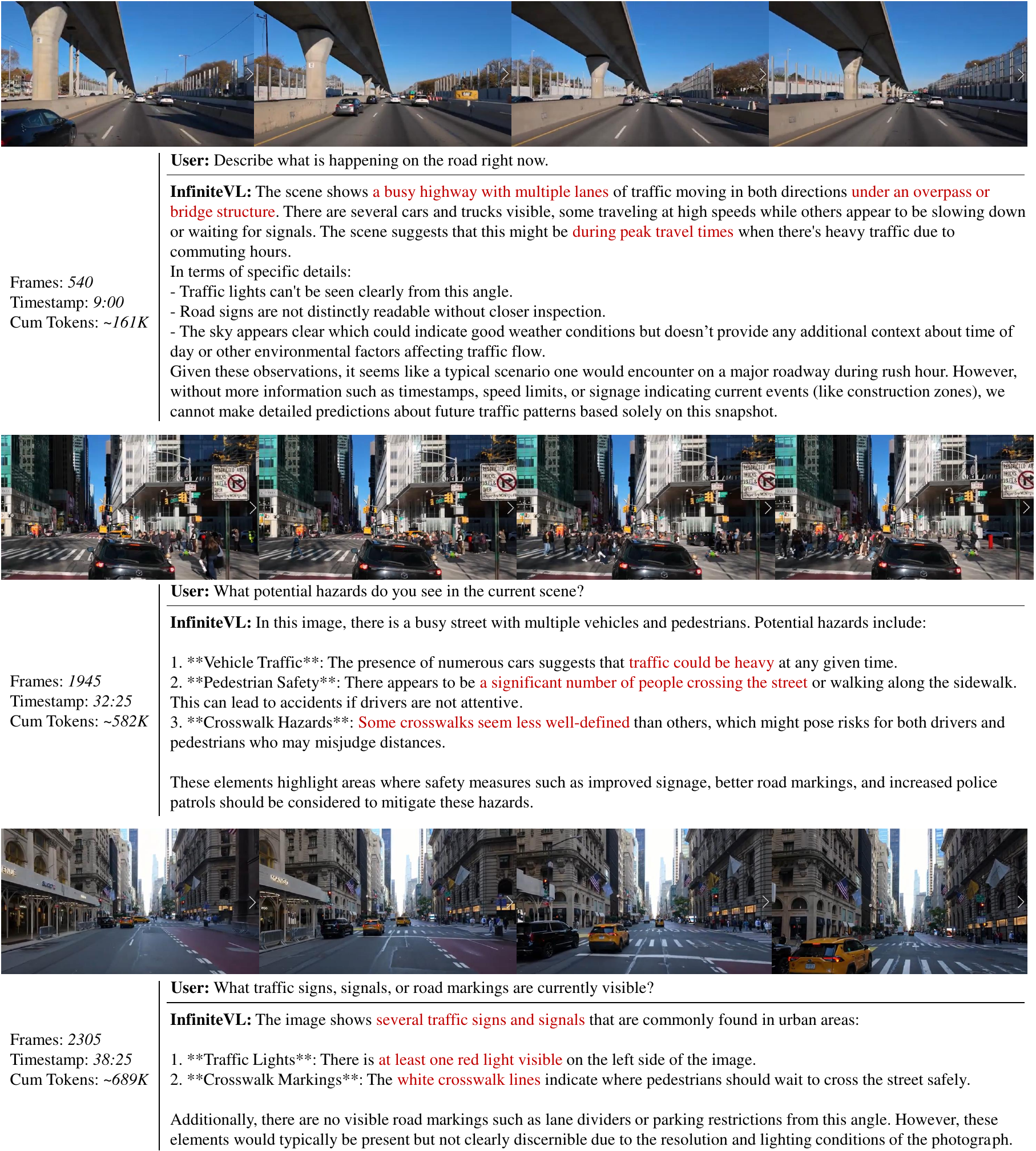}
    \vspace{-0.7cm}
    \caption{Examples of Long-Term Streaming Understanding Capability}
    \label{fig: streaming_case2}
    \vspace{-0.7cm}
\end{figure*}

We present several examples to demonstrate the capabilities of InfiniteVL. As shown in Figure~\ref{fig: image_case1}, InfiniteVL exhibits strong comprehension across high-resolution, text-intensive, and complex structural scenarios. Beyond fundamental vision-language understanding, we further investigate the model's performance in long-term streaming scenarios. Using two extended videos from street-view recordings and dashcam footage, we sample frames at 1 FPS and continuously feed them to InfiniteVL to update its memory cache, while posing questions at randomly selected timestamps. As shown in Figures~\ref{fig: streaming_case1} and~\ref{fig: streaming_case2}, InfiniteVL maintains stable comprehension and reasoning capabilities even under extremely long contexts exceeding 512K tokens.

\end{document}